%% file: main.tex
\documentclass[10pt,twocolumn,letterpaper]{article}

\usepackage[pagenumbers]{cvpr} %

\usepackage{graphicx}
\usepackage{amsmath}
\usepackage{amssymb}
\usepackage{booktabs}
\usepackage{eucal}

\usepackage[pagebackref,breaklinks,colorlinks]{hyperref}
\hypersetup{
    colorlinks=true,
    citecolor=blue,
    }

\usepackage[capitalize]{cleveref}
\crefname{section}{Sec.}{Secs.}
\Crefname{section}{Section}{Sections}
\Crefname{table}{Table}{Tables}
\crefname{table}{Tab.}{Tabs.}

\def\cvprPaperID{***} %
\def\confName{CVPR}
\def\confYear{2022}

\begin{document}

\title{Spatial-Separated Curve Rendering Network for Efficient and High-Resolution Image Harmonization}

\author{
Jingtang Liang\textsuperscript{\rm 1}\thanks{These authors contributed equally to this work},~
Xiaodong Cun\textsuperscript{\rm 1,2}\footnotemark[1],~
Chi-Man Pun\textsuperscript{\rm 1}\thanks{Corresponding author},~
Jue Wang\textsuperscript{\rm 2}\\ 
\textsuperscript{\rm 1}~University of Macau, 
\textsuperscript{\rm 2}~Tencent AI Lab\\
\small\texttt{mb95464@umac.mo, vinthony@gmail.com, cmpun@umac.mo, arphid@gmail.com} 
}

\maketitle

\input{src/tex/abstract}

\input{src/tex/introduction}

\input{src/tex/related_works}

\input{src/tex/methods}

\input{src/tex/exp}

\input{src/tex/conclusion}

{\small
\bibliographystyle{ieee_fullname}
\bibliography{ref}
}

\input{supp_arxiv}

\end{document}

%% file: src/tex/abstract.tex
\begin{abstract}
Image harmonization aims to modify the color of the composited region with respect to the specific background. Previous works model this task as a pixel-wise image-to-image translation using UNet family structures. However, the model size and computational cost limit the ability of their models on edge devices and higher-resolution images. 
To this end, we propose a novel spatial-separated curve rendering network~(S$^2$CRNet) for efficient and high-resolution image harmonization for the first time. 
In S$^2$CRNet, we firstly extract the spatial-separated embeddings from the thumbnails of the masked foreground and background individually. Then, we design a curve rendering module~(CRM), which learns and combines the spatial-specific knowledge using linear layers to generate the parameters of the piece-wise curve mapping in the foreground region.
Finally, we directly render the original high-resolution images using the learned color curve.
Besides, we also make two extensions of the proposed framework via the Cascaded-CRM and Semantic-CRM for cascaded refinement and semantic guidance, respectively. 
Experiments show that the proposed method reduces more than 90\% parameters compared with previous methods but still achieves the state-of-the-art performance on both synthesized iHarmony4 and real-world DIH test set. Moreover, our method can work smoothly on higher resolution images~(\eg, $2048\times2048$) in 0.1 second with much lower GPU computational resources than all existing methods. The code will be made available at: \url{http://github.com/stefanLeong/S2CRNet}.
\end{abstract}

%% file: src/tex/introduction.tex
\section{Introduction}
\vspace{-0.5em}
Image composition~(or image splicing in multimedia security) is a popular and necessary tool for image editing. However, in addition to the serrated edges caused by the irregular borders, the ``style" disharmony occurs when we directly copy source regions~(foreground) to the host image~(background). The disharmony will degrade the quality of the composited images, which also can be distinguished by the human eyes easily. In general, handling this gap requires the professional editing of the well-knowledged experts. Thus, the task of image harmonization aims to squeeze this gap by leveraging some advanced algorithms, which also has a broad impact on image editing, relighting and augmented reality~\cite{relighting,lee2019inserting}.

\begin{figure}[t]

    \centering 
    \captionsetup{justification=centering}
    \captionsetup[subfigure]{labelformat=empty}
    \begin{subfigure}[b]{\columnwidth}
        \centering
         \includegraphics[width=0.8\columnwidth]{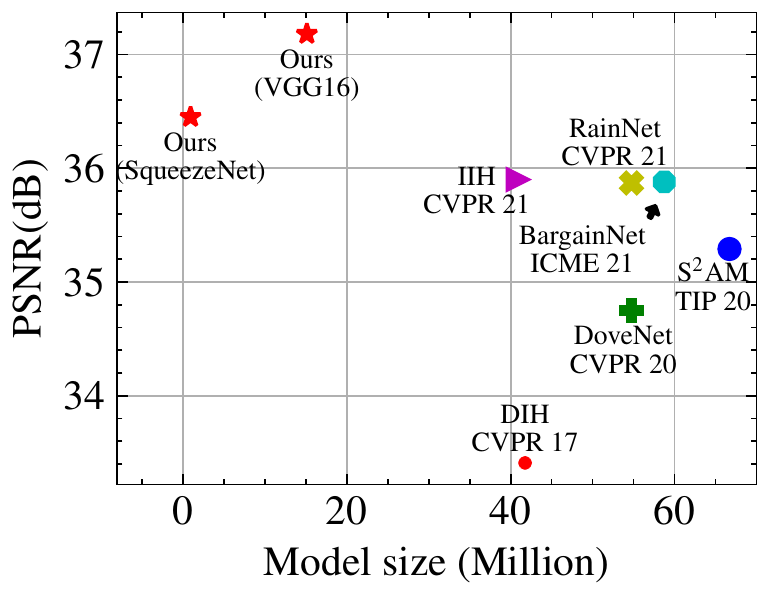}
    \end{subfigure}
    \begin{subfigure}[b]{0.325\columnwidth}
         \centering
         \includegraphics[width=\columnwidth,height=0.7\columnwidth]{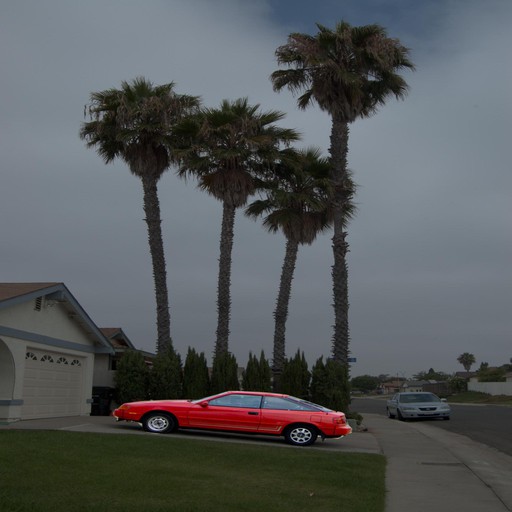}
         \caption{\scriptsize Input~/~Speed~/~MACS}
    \end{subfigure}
    \hfill
    \begin{subfigure}[b]{0.325\columnwidth}
         \centering
         \includegraphics[width=\columnwidth,height=0.7\columnwidth]{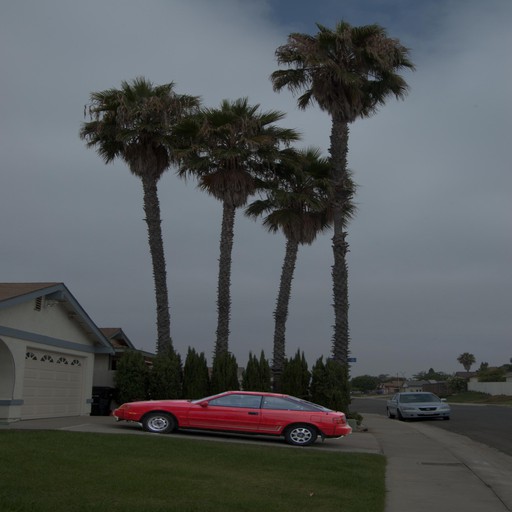}
         \caption{\scriptsize DoveNet~/~0.12~s~/~1.21~T}
    \end{subfigure}
    \hfill
    \begin{subfigure}[b]{0.325\columnwidth}
         \centering
         \includegraphics[width=\columnwidth,height=0.7\columnwidth]{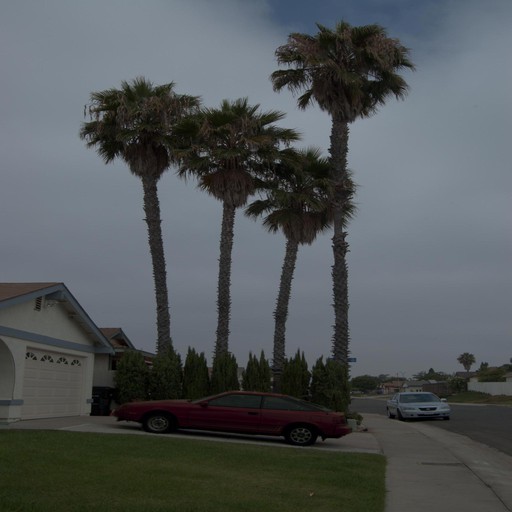}
         \caption{\scriptsize S$^2$AM~/~0.46~s~/~3.02~T}    
    \end{subfigure}
    
    \par\bigskip\vspace*{-1em}
    \begin{subfigure}[t]{0.325\columnwidth}
         \centering
         \includegraphics[width=\columnwidth,height=0.7\columnwidth]{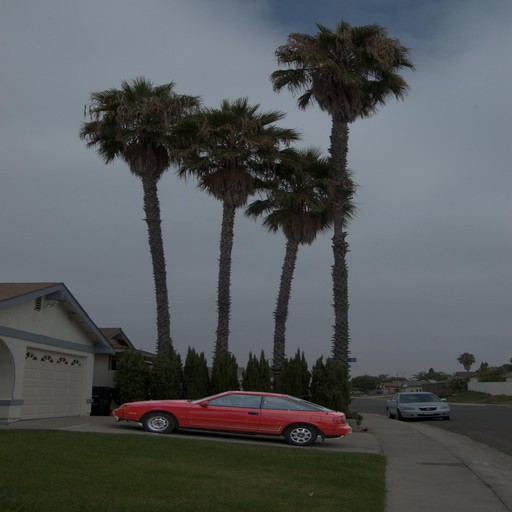}
         \caption{\scriptsize BargainNet/~0.45 s~/~1.23~T}
    \end{subfigure}
    \hfill
    \begin{subfigure}[t]{0.325\columnwidth}
         \centering
         \includegraphics[width=\columnwidth,height=0.7\columnwidth]{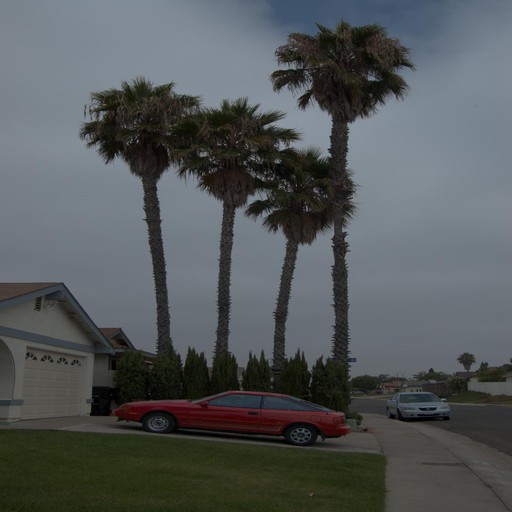}
         \caption{\scriptsize \textbf{Ours}~/~\textbf{0.11~s}~/~\textbf{0.61~G}}
    \end{subfigure}
    \hfill
    \begin{subfigure}[t]{0.325\columnwidth}
         \centering
         \includegraphics[width=\columnwidth,height=0.7\columnwidth]{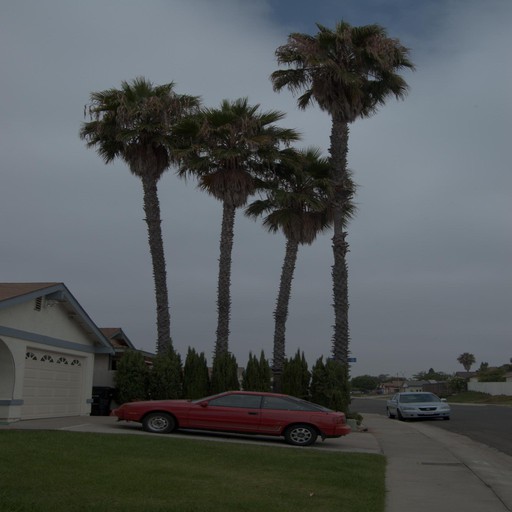}
         \caption{\scriptsize Target Image}
    \end{subfigure}
    \captionsetup{justification=justified}
    \caption{In the top figure, the proposed methods outperform other methods using much less parameters under the same setting~(testing in $256 \times 256$ resolution). Also, given a high-resolution image~(originally $2048 \times 2048$ in this example), our method shows much better performance, lower computational cost~(MACS) and faster speed than previous methods.}
    \vspace{-1em}
    \label{fig:page1}
\end{figure}

Traditional image harmonization methods intend to manually adjust and modify the specific features in the composite images, such as color~\cite{intro_color1,intro_color2}, illumination~\cite{intro_illumination} and texture~\cite{intro_texture},~\emph{etc.}. However, the hand-crafted and statistic low-level features cannot work well for the diverse composite images in complicated real world. Since the deep convolutional neural network~(CNN) has reached impressive performance in many computer vision tasks, several attempts have also been made to address image harmonization tasks. For example, the  semantic clues~\cite{DIH_Tsai, sofiiuk2021foreground}, the spatial differences of the neural network~\cite{s2am, Hao2020bmcv} and generative adversarial network~(GAN~\cite{gan}) based methods~\cite{dovenet,bargainet} have been proposed following the encoder-decoder based structures~(UNet~\cite{unet,pix2pix}) for pixel-wise prediction. Thus, as shown in Figure~\ref{fig:page1}, the speed and computational cost are sensitive to image resolution because those structures require to predict the pixel-wise results. Besides, their model sizes are too large for the edge devices, such as mobile phone. The problems mentioned above restrict the applying range of their methods since the real-world images editing are at any resolution.

\begin{figure}[t!]
    \centering 
    \begin{subfigure}[b]{0.45\columnwidth}
         \centering
         \includegraphics[width=\columnwidth]{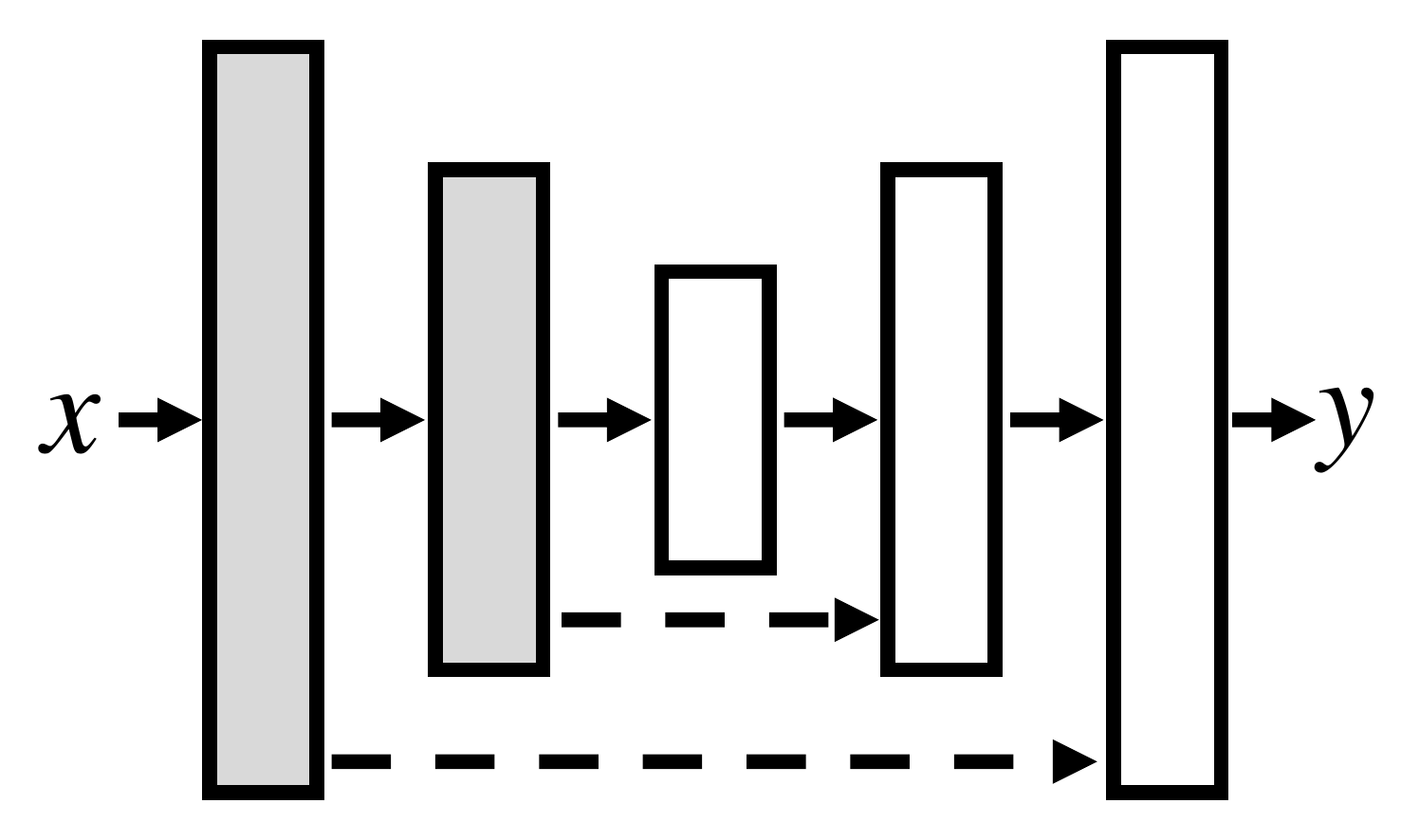}
         \caption{Previous methods}
         \label{fig:unet}
    \end{subfigure}
    \begin{subfigure}[b]{0.45\columnwidth}
         \centering
         \includegraphics[width=\columnwidth]{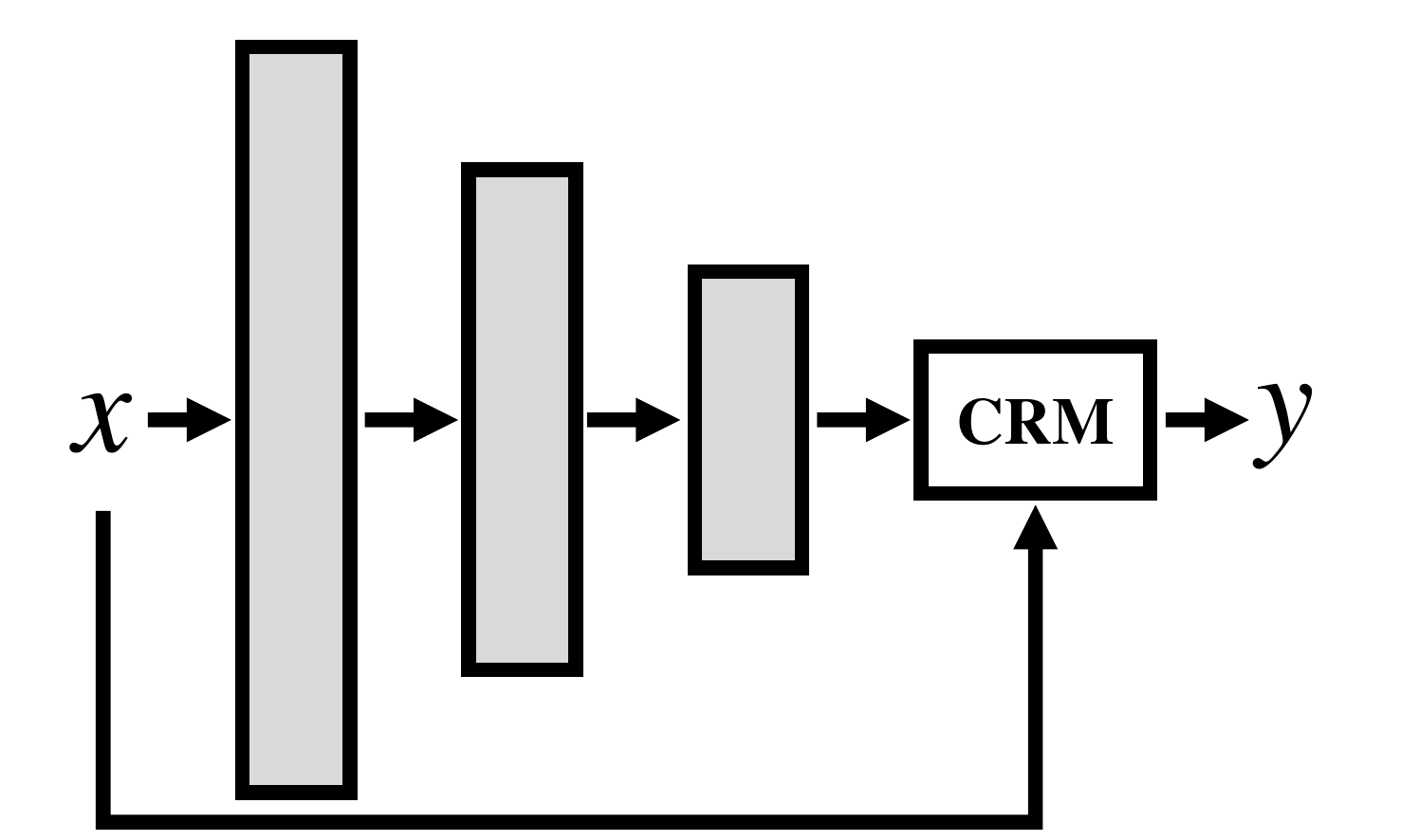}
         \caption{Ours}
         \label{fig:our_backbone}
    \end{subfigure}
    \caption{We learn global mappings for image harmonization and are totally different from previous methods~\cite{s2am,dovenet,bargainet,sofiiuk2021foreground,Hao2020bmcv,DIH_Tsai,Guo_2021_CVPR,Ling_2021_CVPR} that consider it as a pixel-wise image-to-image translation task.}
    \vspace{-1em}
    \label{fig:backbone}
\end{figure}

Differently, in this paper, we rethink the image harmonization in a totally different way: 
Reviewing the image harmonization process in image editing software~($e.g.$ PhotoShop), experts tend to adjust the global properties~(color curve, illuminant, \emph{etc.}) over the whole images rather than the pixel-wise color adjustment. Thus, the global editing can be enabled by considering the properties as the mapping function of the pixels intensities. 
Moreover, this global adjustment is reliably efficient in any resolution images without extra expense of computational cost.

To make the proposed method efficient on images at any resolutions, 
we model the observations above by learning parameters of the \emph{global editing curves} of the composite foreground.
Hence, a novel curve rendering module~(CRM) is designed to produce the image-adaptive parameters of the curves that we will use to render the composite image. Specifically, we first separate the composite image into foreground/background regions using the given foreground mask. Then, we extract the global high-level features from both regions by a shared pre-trained general feature extractor~(SqueezeNet~\cite{iandola2016squeezenet} or VGG16~\cite{simonyan2014very}). Particularly in CRM, the extracted features from foreground/background are learnt by a single layer linear projection for each region separately. Finally, the combination of these two spatial-specific features will be represented as the parameters of color curves, and we render the original foreground for each color channel with the approximate curves we learned. As shown in Figure~\ref{fig:backbone}, the proposed framework works in a totally different way compared with previous methods, which allow our method to run on the higher-resolution and edit image in real time.

Furthermore, we also make two extensions to the proposed framework. On one hand, we propose \emph{semantic}-CRM. Since different foregrounds represent different categories, we learn the class-aware feature embeddings for each category individually by the user-guided foreground semantic encoding.
On the other hand, we propose the \emph{cascaded}-CRM, which is also inspired by the photo editing software since the image editing process generally contains multiple steps. In our implementation, we predict different domain embedding to achieve this goal via a cascaded prediction. 
Benefit by the proposed framework, our method shows a significantly better performance than previous state-of-the-art image harmonization networks with only 2\%~(25\% using VGG16 backbone) of the parameters. Besides, our method can also run much faster than most previous methods with few computation cost on high-resolution images.

\if

\fi

Our main contributions are summarized as follows:

\begin{itemize}
    \item We propose a novel spatial-separated curve rendering network~(S$^2$CRNet). The \textbf{first} network for both efficient and high-resolution image harmonization.
    \item We show the extension ability of the proposed S$^2$CRNet via better backbones or enhanced curve rendering module~(CRM) via the Cascaded-CRM and Semantic-CRM.
    \item Experiments show that our method can achieve state-of-the-art performance and run much faster than the previous methods, while using fewer parameters and lower computational cost. 
\end{itemize}

%% file: src/tex/related_works.tex
\section{Related Works}
\vspace{-0.5em}
\label{sec:related}
\noindent\textbf{Image Harmonization.} Traditional image harmonization  methods aim at improving composite images via low-level appearance features, such as manually adjusting global color distributions \cite{rw_color1,rw_color2}, applying gradient domain composition \cite{rw_gradient1,rw_gradient2} or manipulating multi-scale transformation and statistical analysis \cite{intro_texture}. Although these methods achieve preliminary results in harmonization tasks, the realism of the composite images cannot be visually guaranteed.

\begin{figure*}[t]
    \centering
    \includegraphics[width=\textwidth]{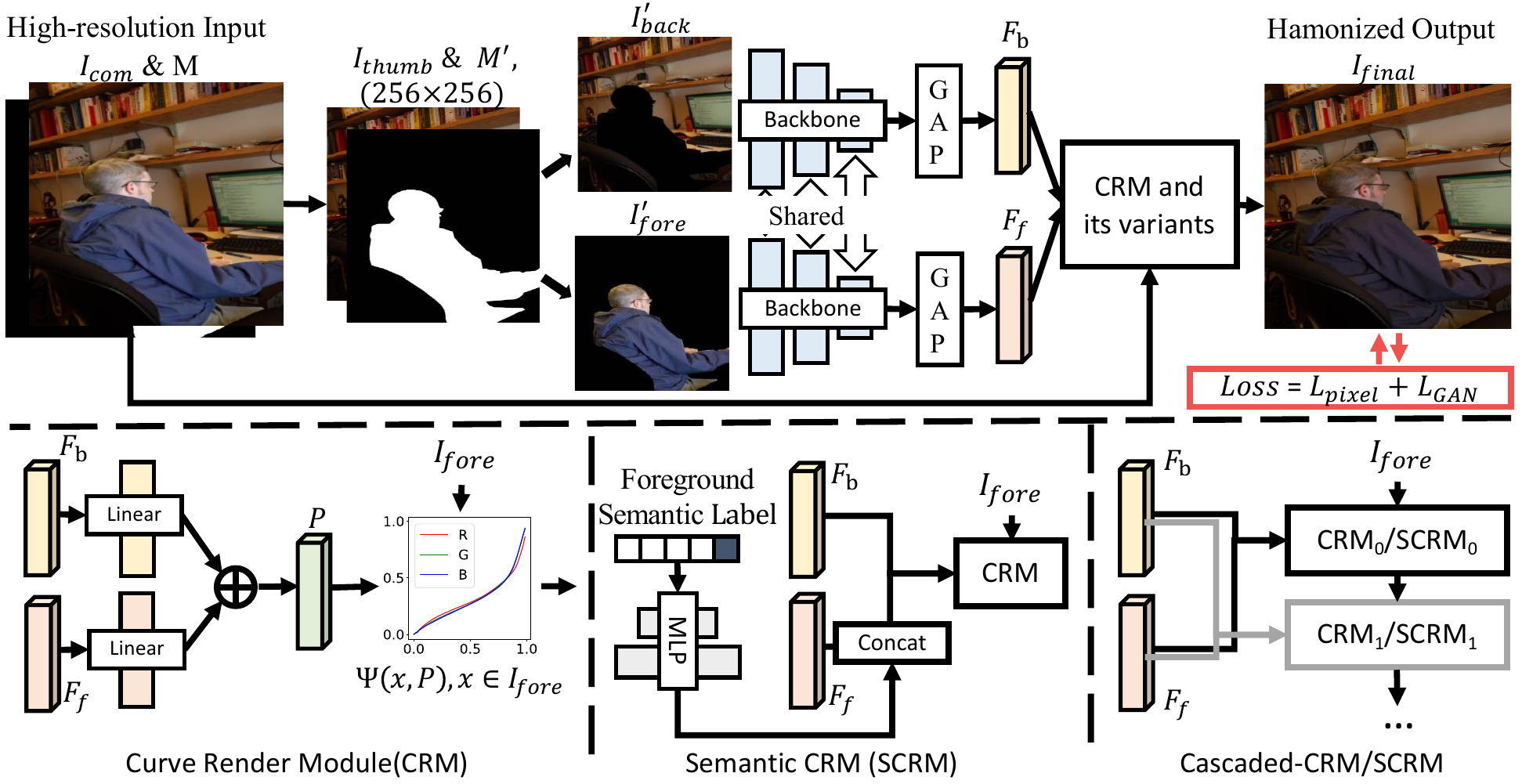}
    \vspace{-2em}
    \caption{The overview of the S$^2$CRNet, including CRM and its two variants: SCRM and Cascaded-CRM/SCRM.}
    \vspace{-1em}
    \label{fig:nn}
\end{figure*}

As the deep learning approaches has been successfully applied to the computer vision tasks, \cite{zhu2015learning} back-propagate a pre-trained visual discriminator model to change the appearance harmony of the composite images. Later, further researches consider this task as an image to image translation problem. For example, additional semantic decoder~\cite{DIH_Tsai} and pre-trained semantic feature~\cite{sofiiuk2021foreground} are used to ensure the semantic consistence between the composite inputs and harmonized outputs. Another noticeable idea is to model the differences between the foreground and background with the given mask. For example, novel spatial-separated attention module~\cite{s2am,Hao2020bmcv} under image-to-image translation framework; Domain-guided features as the discriminator of GAN~\cite{dovenet} and as additional input~\cite{bargainet}; masked-guided spatial normalizations~\cite{Ling_2021_CVPR, Guo_2021_CVPR} for the foreground and background respectively. However, all the previous deep networks still model this task as a pixel-wise image to image translation problem using an encoder-decoder structure, which suffers from computational inefficiency and may degrade the performance and visual quality on high-resolution inputs.

\noindent\textbf{Efficient Network Design for Image Enhancement.}
Efficient networks designed for edge devices have also been widely-discussed in computer vision tasks~\cite{howard2017mobilenets}.
For image enhancement, \cite{Gharbi} introduce the deep bilateral learning for high-resolution and real-time image processing on mobile devices. Also, learning the image-adaptive global style features shows promising results in Exposure~\cite{hu2018exposure} , CURL~\cite{moran2021curl} and  3DLUT~\cite{3dlut} for global image enhancement. Besides, Guo~\etal~\cite{guo2020zero} design a high-order pixel-wise curve function for low-light enhancement.
Since our image harmonization task can be considered as a \emph{regional} image enhancement problem, it is natural to leverage the style curve to image harmonization tasks. However, different from the networks for image enhancement~\cite{hu2018exposure,moran2021curl,3dlut} and low-light enhancement~\cite{guo2020zero}, image harmonization methods rely on regional modification under the guidance of the background. Thus, we design the network structure and learn global mapping functions on this task for the first time.

%% file: src/tex/methods.tex
\vspace{-0.5em}
\section{Method}
\vspace{-0.5em}
\label{sec:methods}
We first show the overall network structure of the proposed method. Then, we give the details of Curve Rendering Module~(CRM) and its variants, which are the key components in S$^2$CRNet, including CRM, \emph{Semantic}-CRM~(SCRM) and their cascaded extensions. Finally, we discuss the loss functions.

\subsection{Overall Network Structure}
\label{sec:overall_network_structure}
 As shown in Figure~\ref{fig:nn}, given a high-resolution composite image $ I_{com} \in \mathbb{R}^{3 \times H \times W} $ and its binary mask $ M \in \mathbb{R}^{1 \times H \times W} $ of the corresponding foreground,  we first get the thumbnail image $I_{thumb} \in \mathbb{R}^{3 \times h \times w} $ and the mask $ M' \in \mathbb{R}^{1 \times h \times w} $ by down-sampling $I_{com}$ and $M$ with a factor of $H/h$ for fast inference and the minimal computational cost. 
For the spatial-separated feature encoding, we first segment the thumbnail image $I_{thumb}$ into foreground and background via the mask $M'$ and inverse mask $M'_{inv} = 1-M'$, respectively. 
Next, given the foreground $I_{fore} = I_{thumb} \times M'$ and background $I_{back} = I_{thumb} \times M'_{inv}$ images, we use a shared domain encoder $\Phi$ to extract the spatial-separated features for foreground and background respectively. 
Here, we choose the SqueezeNet~\cite{iandola2016squeezenet} as the domain encoder~(backbone), which is pre-trained on the ImageNet~\cite{imagenet} and we only use the first 12 layers to get deeper color feature embedding. 
We also try different backbones~(\eg, VGGNet~\cite{simonyan2014very}) to achieve better performance as shown in Table~\ref{table:psnr}. While considering the purpose of this paper is for efficient and high-resolution image harmonization, thus we use SqueezeNet as our default backbone for its good balance between the efficiency and effectiveness.

After obtaining the embedding foreground $F_{fore} \in \mathbb{R}^{D \times h' \times w'} $ and background $ F_{back} \in \mathbb{R}^{D \times h' \times w'}$ features from the domain encoder, we squeeze the foreground/background feature dimensionally via the global average pooling to avoid the influence of spatial information.
Then, foreground $F_f \in \mathbb{R}^{D}$ and background $F_b \in \mathbb{R}^{D}$ are learnt to generate the parameters of the color curve and render the channel-wise color curve via the proposed \emph{Curve Rendering Module} automatically. We will discuss the details and its variants in the later sections.

\subsection{Curve Rendering Modules and its Variants}
\label{sec:crm}
We first introduce the basic idea behind the proposed network via the \emph{Curve Rendering Module}~(CRM). Then, we discuss two different extensions using the semantic label and recurrent refinement.

\textbf{Curve Rendering Module~(CRM).}
Most previous image harmonization methods~\cite{dovenet,s2am,sofiiuk2021foreground,Hao2020bmcv} consider this task as a pixel-wise image to image translation task, which is heavy and only works on certain resolution as we discussed in the related works. Differently, we model this task as a  global region image enhancement task. Thus, our goal of CRM is try to adjust the foreground color under the given background.

To achieve the above goal, as shown in Figure~\ref{fig:nn}, after obtaining the spatial-separated foreground embedding $F_{f}$ and background embedding $F_{b}$ from the domain encoder separately, we first embed these spatial-aware features using two projection functions $\phi_{f}(\cdot)$/$\phi_{b}(\cdot)$ for foreground/background correspondingly, where each projection function is a single linear layer with ReLU activation. Then, to harmonize the foreground under the guidance of the related background features, we get $P \in \mathbb{R}^{3L}$ by performing channel-wise addition between $\phi_{f}(F_{f})$ and $\phi_{b}(F_{b})$. Here, $L$ includes the parameters of R, G, B color channels and each channel has $L=64$ piece parameters for the balance between the computational complexity and performance. 

\begin{figure}[t!]
    \centering
    \includegraphics[width=0.9\columnwidth]{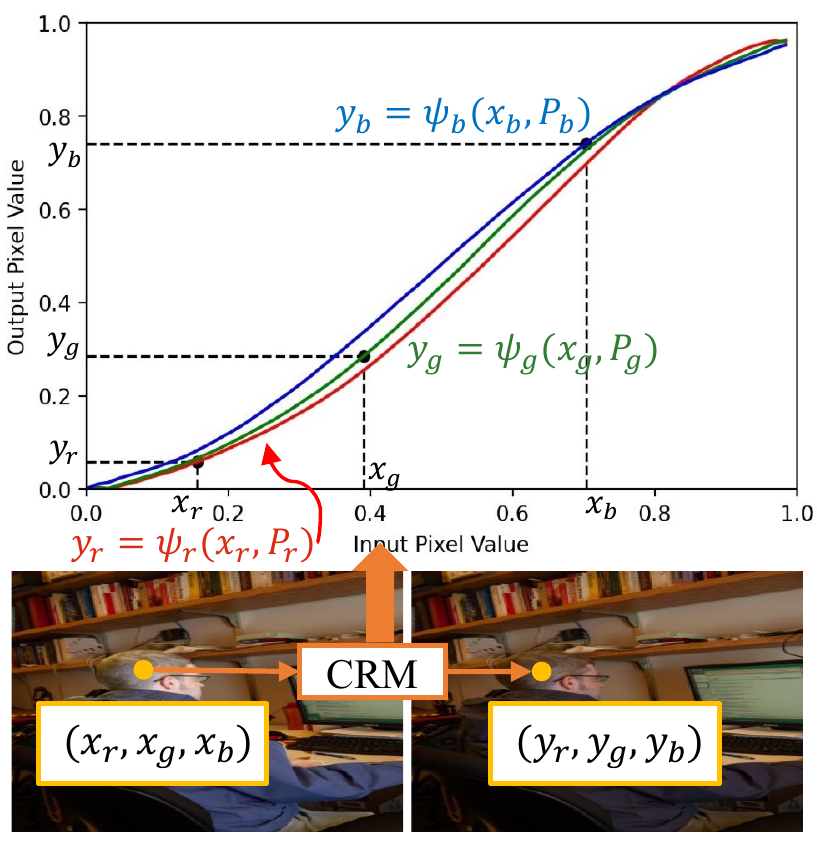}
    \vspace{-1em}
    \caption{CRM maps the input pixels to the target pixels using curve function $\psi(\cdot)$, where the parameters $P$ of $\psi(\cdot)$ are learnt from the embeddings of the spatial-aware encoders.}
    \vspace{-1em}
    \label{fig:curve}
\end{figure}

Since this hybrid feature $P$ contains both the information from the background and foreground, it can be a good representation for the guidance of the foreground editing. 
To better modeling the color-wise changes, we consider the mappings between intensities rather than the semantic information. Thus, we choose the color curve as the editing tool and make it differentiable~\cite{hu2018exposure} by approximating $L$ levels monotonous piece-wise linear function, and then rendering the original pixels in the foreground region. As shown in Figure~\ref{fig:curve}, for each pixels~($x_r, x_g, x_b$) in the foreground of the original composited image, we use CRM to map it with the learnt color curve. 
Here, the mappings of each intensity is identical and not related to the specific location or semantic information. The parameters of the piece-wise linear function is provided and learnt by the spatial-separated encoder and each channel is learnt individually.

Mathematically, after getting the mixed embedding for each channel $P^c = [p_{0}, p_{1}, p_{2}, \ldots, p_{L-1}]$, we render the foreground $I_{fore}^{c}$ ($ c \in \{R,G,B\}$) of the composite image via the curve rendering function $\psi(I_{fore}^{c}, P^{c})$, which can be denoted as:
\begin{equation}
\begin{aligned}
\nonumber \psi_{c}(I_{fore}^{c}, P_{c})&=\frac{1}{\sum_{j=0}^{L-1} p_{j}} \sum_{i=0}^{L-1} p_{i}\xi\left(x -\frac{i}{L}\right), x \in I_{fore}^{c},\\
\end{aligned}
\end{equation}
\begin{equation}
\begin{aligned}
\text{where} \quad \xi(y)&=\left\{
\begin{array}{lc}
    0,\quad y<0 \\
    y,\quad 0 \leq y<1 \\
    1,\quad y>1
    \end{array}  \right.
\end{aligned}
\label{eq:cr}
\end{equation}

Finally, the harmonized image can be obtained by the combination of the original background: $I_{final} = \Psi(I_{fore}, P) + I_{back}$.

\textbf{Semantic CRM.}
\label{sec:SCRM}
Previous methods~\cite{dovenet,s2am} intend to obtain a unified harmonization model for any foreground images without any specific semantic knowledge. However, the semantic information is also important for the image harmonization~\cite{DIH_Tsai,sofiiuk2021foreground} and it does not make sense if we apply the same style to harmonize different categories~(\eg $Car$ and $Person$).
Since we have supposed that the linear layers in the CRM contain the domain knowledge of the foreground, we make a further step by adding extra semantic label of the foreground object to our vanilla CRM.

As shown in Figure~\ref{fig:nn}, given the semantic label $d$ of the foreground region, we first embed the labels using a two-layer Multi-layer Perceptron~(MLP), obtaining the semantic-aware embedding $D$. Then, we concatenate the embedded feature from the network $\Phi$ and the label embedding $D$ to the CRM. For semantic label definition, we analyze the categories of the foreground regions in iHarmony4 and divide it into 5 classes as guidance, including \emph{Person}, \emph{Vehicle}, \emph{Animal}, \emph{Food} and others. More details can be found in the supplementary materials.

\textbf{Cascaded CRM/SCRM.}
It is natural for the image editing tools to adjust the images with multiple steps for better visual quality. Inspired by this phenomenon, we extend our CRM~(or SCRM) via the cascaded refinements. To reduce the inference time and learn a compact model, we keep the global features from the backbone unchanged and generate multi-stage heads and give the supervisions of each stage.

As shown in Figure~\ref{fig:nn}, given the global foreground features $F_f$ and background features $F_b$ from the backbone, we firstly generate $P_{0}$ via a CRM and get its rendered image $I_{0}$ using $\Psi_{c}(I_{fore}^{c}, P_{0})$. Then, we use another set of linear layers to predict the parameters $P_{n}$ from the same global features~($F_f$, $F_b$) and rendering the curve using the previous prediction $I_{n-1}$ via $\Psi_{c}(I_{n-1}, P_{n})$.
We set $n$ equals to 2 to ensure the fast inference as efficiency as well as the high harmonization quality.

\newcommand{\imname}[2]{src/figures/Har/#1/#2}
\newcommand{\methods}{input,DoveNet,BargainNet,s2am,Ours_s,Ours_v,Target}
\newcommand{\imgs}{a2999_1_5,f159_1_2,f1757_1_2}

\begin{figure*}[tbh!]
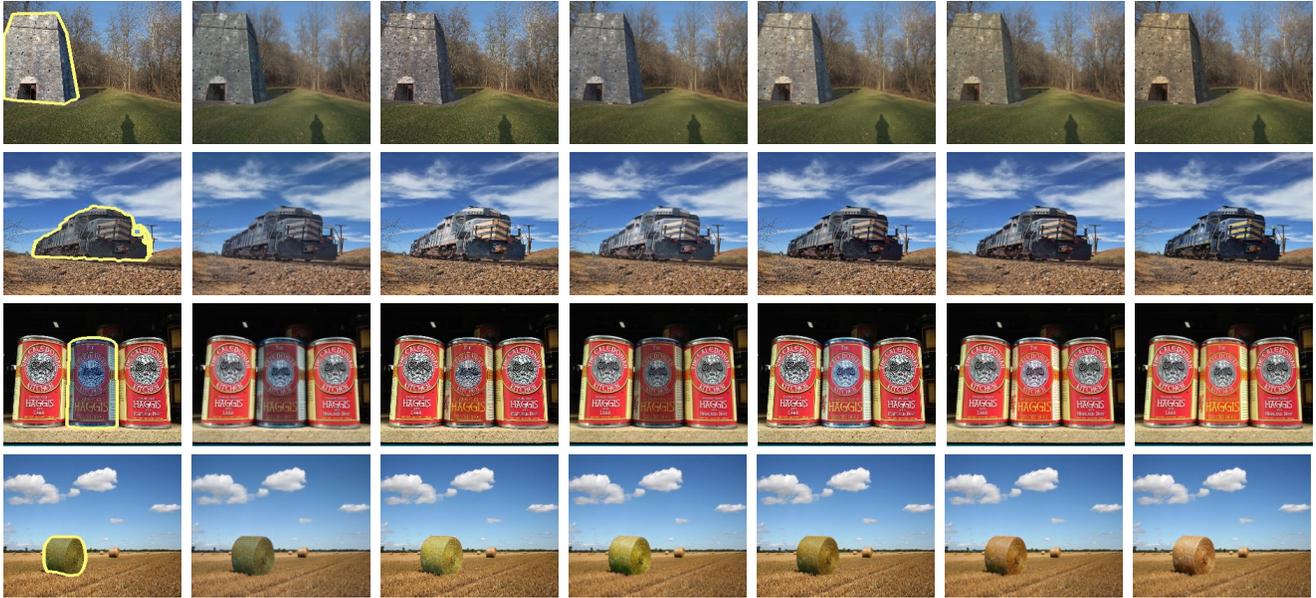

\centering     %
\makeatletter
\@for\img:=\imgs\do{
    \@for\method:=\methods\do{
    \begin{subfigure}[t]{0.135\textwidth}
            \centering
     \includegraphics[width=\columnwidth,height=0.8\columnwidth]{\imname{\method}{\img}}
    \end{subfigure}\hfill
    }
    \vfill\vspace{0.25em}
}
\begin{subfigure}[t]{0.135\textwidth}\centering
     \includegraphics[width=\columnwidth,height=0.8\columnwidth]{\imname{input}{f1774_1_1}}\caption{Input}
\end{subfigure}\hfill
\begin{subfigure}[t]{0.135\textwidth}\centering
     \includegraphics[width=\columnwidth,height=0.8\columnwidth]{\imname{DoveNet}{f1774_1_1}}\caption{DoveNet}
\end{subfigure}\hfill
\begin{subfigure}[t]{0.135\textwidth}\centering
     \includegraphics[width=\columnwidth,height=0.8\columnwidth]{\imname{BargainNet}{f1774_1_1}}\caption{BargainNet}
\end{subfigure}\hfill
\begin{subfigure}[t]{0.135\textwidth}\centering
     \includegraphics[width=\columnwidth,height=0.8\columnwidth]{\imname{s2am}{f1774_1_1}}\caption{S$^2$AM}
\end{subfigure}\hfill
\begin{subfigure}[t]{0.135\textwidth}\centering
     \includegraphics[width=\columnwidth,height=0.8\columnwidth]{\imname{Ours_s}{f1774_1_1}}\caption{S$^2$CRNet-S}
\end{subfigure}\hfill
\begin{subfigure}[t]{0.135\textwidth}\centering
     \includegraphics[width=\columnwidth,height=0.8\columnwidth]{\imname{Ours_v}{f1774_1_1}}\caption{S$^2$CRNet-V}
\end{subfigure}\hfill
\begin{subfigure}[t]{0.135\textwidth}\centering
     \includegraphics[width=\columnwidth,height=0.8\columnwidth]{\imname{Target}{f1774_1_1}}\caption{Target}\hfill
\end{subfigure}
\makeatother
\vspace{-2em}
\caption{Comparisons with other methods on iHarmony4 Dataset. Here, we visualize the input mask as yellow marked region for easy reading. S$^2$CRNet-S and S$^2$CRNet-V denote our method employs SqueezedNet and VGG16 backbone, respectively.}
\label{fig:comp}
\end{figure*}

\subsection{Loss Function}
\label{sec:loss}
We consider image harmonization as a supervised problem. Specifically, we measure the difference between the target and the corresponding rendered images~(for each stage) in the composited region. Thus, we use relative $L_1$ loss between the predicted foreground and the target via the foreground mask $M$. Besides, for better visual quality, we also leverage the adversarial loss~\cite{gan} to our framework. We give the details of each part as follows.

\textbf{Relative $L_1$ Loss $L_{pixel}$.}
Another key idea to make our method work is that
we only calculate the metric between the foreground of the predicted image and the target, where the differences are only measured in a single domain. Thus, inspired by recent works in watermark removal~\cite{bvmr,cun2020split}, we perform the pixel-wise $L_1$ loss in the foreground region $M$ by masking out the background pixels and setting the meaningful region. Specifically, giving the rendered images $I_{n}$ in each stage, we calculate the loss over the masked region: 
\begin{equation}
L_{pixel}=\sum_{n=1}^{N} \frac{\left ||M \times I_{n}-M \times I_{gt}\right ||_{1}}{sum(M)}
\end{equation}
where $N=2$ is the number of iterations.

\textbf{Adversarial Loss $L_{adv}$.}
By considering the proposed S$^2$CRNet as the generator $G$, we also utilize an additional discriminator $D$ to identify the naturalness of the color. In detail, we use a standard 5 layers CONV-BN-RELU discriminator~\cite{cyclegan} and leverage a least squares GAN~\cite{lsgan} as criteria. Then, the generator is learnt to fool the discriminator and the discriminator is trained to identify the real or fake feed images iteratively.

Overall, our algorithm can be trained in an end-to-end function via the combination of the losses above:
\begin{equation}
L_{all}=\lambda_{pixel} L_{pixel}+\lambda_{adv} L_{adv}
\end{equation}
where all the hyper-parameters~($\lambda_{pixel}$ and $\lambda_{adv}$) are empirically set to 1 for all our experiments.

\begin{table*}[t!]
\centering
\resizebox{\textwidth}{!}{%
\begin{tabular}{@{}|c|l|l|cc|cc|cc|cc|cc@{}}
\toprule
\multicolumn{2}{c|}{Sub-dataset}   &   & \multicolumn{2}{c|}{HCOCO}  & \multicolumn{2}{c|}{HAdobe5k}  & \multicolumn{2}{c|}{HFlickr}  & \multicolumn{2}{c|}{Hday2night}  & \multicolumn{2}{c}{All}  \\ \hline
\multicolumn{2}{c|}{Evaluation metric}       & \# Param.         & MSE$\downarrow$ & PSNR$\uparrow$ & MSE$\downarrow$ & PSNR$\uparrow$ & MSE$\downarrow$ & PSNR$\uparrow$ & MSE$\downarrow$ & PSNR$\uparrow$ & MSE$\downarrow$ & PSNR$\uparrow$ \\ \hline
\multicolumn{2}{l|}{Input Composition}         & -              & 67.89                        & 34.07                       & 342.27                        & 28.14                       & 260.98                       & 28.35                       & 107.95                        & 34.01                       & 170.25                        & 31.70                       \\
\multicolumn{2}{l|}{Lalonde \& Efros~\cite{intro_color1}} & -              & 110.10                        & 31.14                       & 158.90                        & 29.66                       & 329.87                       & 26.43                       & 199.93                       & 29.80                        & 150.53                       & 30.16                       \\
\multicolumn{2}{l|}{Xue~\etal~\cite{intro_illumination}}               & -              & 77.04                        & 33.32                       & 274.15                       & 28.79                       & 249.54                       & 28.32                       & 190.51                       & 31.24                       & 155.87                       & 31.40                        \\
\multicolumn{2}{l|}{Zhu~\etal~\cite{zhu2015learning}}               & -              & 79.82                        & 33.04                       & 414.31                       & 27.26                       & 315.42                       & 27.52                       & 136.71                       & 32.32                       & 204.77                       & 30.72                       \\
\multicolumn{2}{l|}{DIH\cite{DIH_Tsai}}                     & 41.76M              & 51.85                        & 34.69                       & 92.65                        & 32.28                       & 163.38                       & 29.55                       & 82.34                        & 34.62                       & 76.77                        & 33.41                       \\

\multicolumn{2}{l|}{DoveNet\cite{dovenet}}                 & 54.76M         & 36.72                        & 35.83                       & 52.32                        & 34.34                       & 133.14                       & 30.21                       & 54.05                        & 35.18                       & 52.36                        & 34.75                       \\
\multicolumn{2}{l|}{S$^2$AM\cite{s2am}}                    & 66.70M         & 33.07 & 36.09 & 48.22 & 35.34 & 124.53 & 31.00 & \underline{48.78} & 35.60 & 48.00 & 35.29         \\
\multicolumn{2}{l|}{BargainNet\cite{bargainet}}              & 58.74M         & \underline{24.84}                        & 37.03                       & \underline{39.94}                        & 35.34                       & \textbf{97.32}  & 31.34                       & 50.98                        & 35.67                       & \underline{37.82}   & 35.88                       \\
\multicolumn{2}{l|}{IIH\cite{Guo_2021_CVPR}}   & 40.86M & {24.92} & 37.16 & 43.02 & 35.20 & {105.13} & 31.34   & 55.53    & 35.96   & {38.71} & 35.90                      \\
\multicolumn{2}{l|}{RainNet\cite{Ling_2021_CVPR}}   & 54.75M &  31.12 & 36.59 & {42.84} & \underline{36.20} & 117.59 & 31.33    & \textbf{47.24}    & 36.12          & 44.50 &35.88                     \\ \hline

\multicolumn{2}{l|}{S$^2$CRNet-SqueezeNet} & \textbf{0.95M} & 28.25 & \underline{37.65} & 44.52  & {35.93} & 115.46 & \underline{31.63}  & 53.33 & \underline{36.28} & 43.20  & \underline{36.45}  \\
\multicolumn{2}{l|}{S$^2$CRNet-VGG16} & \underline{15.14M} & \textbf{23.22} & \textbf{38.48} & \textbf{34.91}  & \textbf{36.42} & \underline{98.73} & \textbf{32.48}  & 51.67 & \textbf{36.81} & \textbf{35.58}  & \textbf{37.18}  \\
\bottomrule
\end{tabular}
}
\vspace{-1em}
\caption{Comparisons on iHarmony4. The best and the second best are marked as boldface and underline respectively. }
\vspace{-1em}
\label{table:psnr}
\end{table*}

%% file: src/tex/exp.tex
\section{Experiments}
\label{sec:exp}

\subsection{Implementation Details}
\label{sec:imp}
We implement our method in Pytorch~\cite{pytorch} and train on a single NVIDIA TITAN V GPU with 12GB memory. The batch size is set to 8 and we train 20 epochs~(50 epochs for VGG16 backbone) for convergence. All the images are resized to 256$\times$256 and random cropped and flipped for fair training and evaluation as previous methods~\cite{dovenet,s2am}. We leverage the AdamW optimizer~\cite{adamw} with the learning rate of $2\times10^{-4}$, the weight decay value of $10^{-2}$ and momentum of 0.9. 

Following previous works~\cite{s2am,DIH_Tsai,dovenet}, we train our model on the benchmark dataset iHarmony4~\cite{dovenet} and test on the test set of iHarmony4 and the real-world test set in DIH~\cite{DIH_Tsai}~(DIH99). The iHarmony4 consists of 4 sub-datasets~(HCOCO, HAdobe5k, HFlickr, Hday2night) and includes 73146 image pairs for image harmonization including the synthesized images, the corresponding foreground masks and the target images. DIH99 only contains 99 real copy-paste samples with its foreground mask. As for evaluation, we validate our approaches on the iHarmony4 using Mean-Square-Errors~(MSE), Peak Signal-to-Noise Ratio (PSNR) and Structural SIMilarity~(SSIM) as criteria metrics. Since DIH99 does not contain the target images, we conduct the subjective experiments.

\subsection{Comparison with Existing Methods}
\label{sec:comparsion}
\vspace{-0.5em}
\textbf{Performance Comparison on iHarmony4.}
We compare our methods with other state-of-the-art image harmonization algorithms, including DoveNet, S$^2$AM, BargainNet, IIH~\cite{Guo_2021_CVPR}, RainNet~\cite{Ling_2021_CVPR}, \emph{etc.}. In our experiments, we choose the Cascaded-SCRM model in different backbones~(SqueezeNet and VGG16 as shown in Table~\ref{table:psnr}), where the semantic labels are generated by a pre-trained segmentation model~\cite{zhou2018semantic}). All previous methods are tested using their official implementations and pre-trained models for fair comparison. As shown in Table~\ref{table:psnr}, even training and testing on 256$\times$256 limits the high-resolution performance, our S$^2$CRNet-SqueezeNet only use 2\% of the parameters to achieve the state-of-the-art performance in PSNR metric, which demonstrates the effectiveness of the proposed network. On the other hand, when using VGG16 backbone~(S$^2$CRNet-VGG16), our method 
outperforms other related methods by a clear margin and still uses only 40\% of the parameters. Moreover, the proposed method also works better even on higher-resolution images, which will be discussed in later section.  

Besides the numeric comparison, our proposed method also obtains better visual quality than others. Qualitative examples in Figure~\ref{fig:comp} show that the proposed method can generate harmonized results that are more realistic than other methods, which further indicates the benefits of the proposed framework. More visual comparisons are presented in the supplementary materials.

\begin{figure*}[t!]
     \centering
     \begin{subfigure}[b]{0.32\textwidth}
         \centering
         \includegraphics[width=\textwidth]{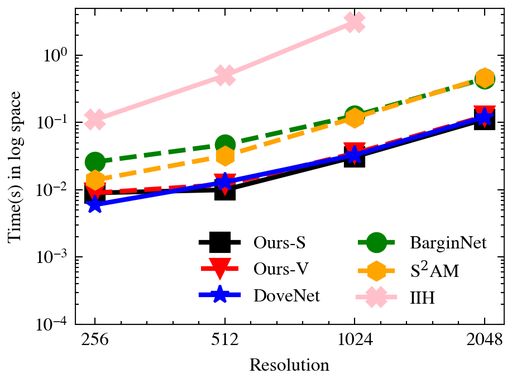}
         \vspace{-1.5em}
         \caption{Running time in \emph{log} scale.}
         \label{fig:hr_speed}
     \end{subfigure}
     \hfill
     \begin{subfigure}[b]{0.32\textwidth}
         \centering
         \includegraphics[width=\textwidth]{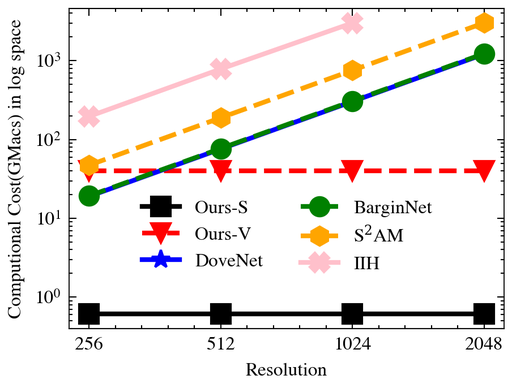}
         \vspace{-1.5em}
         \caption{Computational cost in \emph{log} scale.}
         \label{fig:hr_gmacs}
     \end{subfigure}
     \hfill
     \begin{subfigure}[b]{0.32\textwidth}
         \centering
         \includegraphics[width=\textwidth]{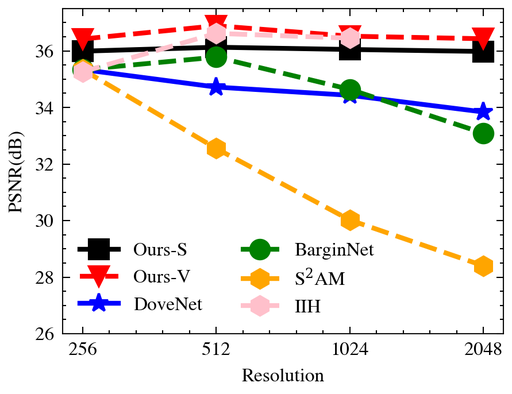}
         \vspace{-1.5em}
         \caption{Performance comparisons.}
         \label{fig:hr_psnr}
     \end{subfigure}
     \vspace{-1em}
    \caption{The influence of the image resolution from different aspects. IIH~\cite{Guo_2021_CVPR} cause out of memory error on $2048\times2048$ images. }
    \vspace{-1em}
    \label{fig:hr}
\end{figure*}

\textbf{High-Resolution Image Harmonization.}
We conduct extra experiments on the HAdobe5k sub-dataset in iHarmony4 to verify the speed and performance of the proposed method on higher-resolution. As experiment setup, we resize the source composite images with the square resolution of 256, 512, 1024 and 2048, and test the average processing time, computational cost and PSNR scores on the same hardware platform.
Since other state-of-the-art methods~(DoveNet, BargainNet, S$^2$AM, IIH) employ the fully convolutional encoder-decoder structures, they can be tested directly in higher resolution. As for our method, we test two backbones of the proposed S$^2$CRNet and donate them as Ours-S~(SqueezeNet as backbone) and Ours-V~(VGG16 as backbone) shown in Figure~\ref{fig:hr}.

As shown in Figure~\ref{fig:hr_speed}, we plot the speed of different image harmonization methods in the \emph{log} space. All the methods suffer a speed degradation with the resolution increasing. However, our the research quality code of Ours-S and Ours-V runs much faster~(0.1s around on a $2048\times2048$ image) than all other methods and is nearly 5$\times$ faster than S$^2$AM and BargainNet. Also, since we use a fixed size input, the required computation cost of our method still much less than previous methods as the resolution increasing as shown in Figure~\ref{fig:hr_gmacs}. 
In terms of the harmonization quality, there are also some interesting phenomenons. As shown in Figure~\ref{fig:hr_psnr}, most of other methods confront a significant performance decline as the resolution increases. It might be because the encoder-decoder based structure will produce different reception fields of original images and then downgrade its performance. Differently, our methods maintain the higher performance at all resolutions.

\textbf{User Study on DIH99.}
Since the real-wold image composition is still different from the synthesized dataset, we evaluate the proposed method~(S$^2$CRNet-SqueezeNet) and existing methods~(DIH, DoveNet, BarginNet) by subjective experiments on DIH99. In detail, we randomly shuffle the displaying order of all the images and invite 18 users to select the most realistic results. As shown in Table~\ref{table:user_study}, the proposed method gets the most votes with faster inference time and fewer model parameters as discussed previously. More details of the user study and more harmonization results of the real composite samples are shown in the supplementary.

\begin{table}[h!]
\centering
\resizebox{\columnwidth}{!}{%
\begin{tabular}{cccccc}
\hline
Method      & Input       & DIH         & DoveNet     & BargainNet  & Ours        \\ \hline
Total votes & 224         & 385         & 403         & 328         & \textbf{442}         \\
Preference   & 12.57$\%$ & 21.60$\%$ & 22.62$\%$ & 18.41$\%$ & \textbf{27.80$\%$} \\ \hline
\end{tabular}
}
\vspace{-1em}
\caption{User study on DIH99 test set.}
\vspace{-1em}
\label{table:user_study}
\end{table}

\subsection{Ablation Studies}
\label{sec:ablation_study}
\vspace{-0.5em}
We conduct the ablation experiments to demonstrate the effectiveness of each component in the proposed S$^2$CRNet. All the experiments are performed on both HCOCO and iHarmony4 with same configurations using the SqueezeNet backbone. 

\begin{table}[tbh!]
\centering
\resizebox{\columnwidth}{!}{%
\begin{tabular}{@{}c|cc|cc|cc|cc@{}}
\toprule
\multicolumn{1}{c|}{\#} &\multicolumn{2}{c|}{Loss} & \multicolumn{2}{c|}{Network}  & \multicolumn{2}{c|}{HCOCO} & \multicolumn{2}{c}{iHarmony} \\
&$L_{pixel}$                  & $L_{adv}$         & $\Phi$ & CRM                & MSE $\downarrow$         & PSNR $\uparrow$       & MSE $\downarrow$          & PSNR $\uparrow$     \\ \hline
- & \multicolumn{4}{c|}{Original Input} & 67.89 & 34.07 & 170.25 & 31.70 \\ \hline
${A}$ & $L_1$   &   & Ours & $\checkmark$ & 67.64 & 34.08 & 114.65 & 32.11 \\
${B}$ & $rL_1$ &  & Ours  &    $\checkmark$    & 28.43        & 37.59       & 46.79         & 36.20        \\
${C}$ & $rL_1$ & $\checkmark$  & Ours  &    $\checkmark$    & 29.45        & 37.51       & 45.17         & 36.27        \\
${D}$ & $rL_1$   &   $\checkmark$  & $I_{fore}$ & $\checkmark$                  & 34.62        & 36.98       &    79.73           &     34.57         \\
${E}$ &  $rL_1$ &  $\checkmark$  & $I_{com}$   &     $\checkmark$                         & 58.53        & 34.69       &    88.61           &      33.88        \\
${F}$ & $rL_1$ &    $\checkmark$  & Ours  &  C  & 28.47        & 37.60        & 44.08         & 36.41        \\
${G}$ & $rL_1$     &  $\checkmark$ & Ours  &  CS & \textbf{27.40}         & \textbf{37.72}       & \textbf{43.20}          & \textbf{36.45}        \\
\bottomrule
\end{tabular}
}
\vspace{-1em}
\caption{Ablation studies.}
\vspace{-1em}
\label{table:ablation}
\end{table}

\begin{figure}[t!]
    \centering 
    \begin{subfigure}[b]{0.242\columnwidth}
         \centering
         \includegraphics[width=\columnwidth]{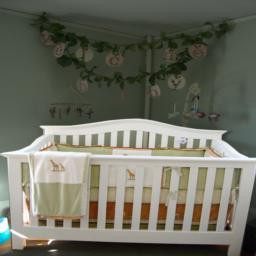}
    \end{subfigure}
    \hfill
    \begin{subfigure}[b]{0.242\columnwidth}
         \centering
         \includegraphics[width=\columnwidth]{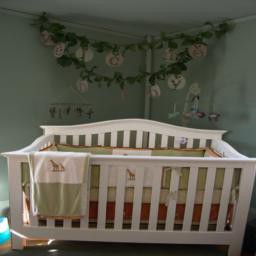}
    \end{subfigure}
    \hfill
    \begin{subfigure}[b]{0.242\columnwidth}
         \centering
         \includegraphics[width=\columnwidth]{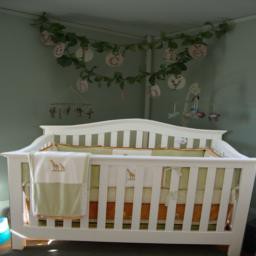}
    \end{subfigure}
    \begin{subfigure}[b]{0.242\columnwidth}
         \centering
         \includegraphics[width=\columnwidth]{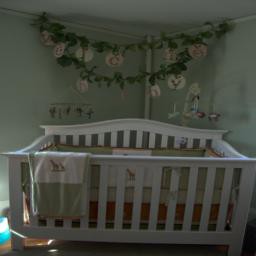}
    \end{subfigure}

    \par\bigskip\vspace*{-1em}
    \begin{subfigure}[t]{0.242\columnwidth}
         \centering
         \includegraphics[width=\columnwidth]{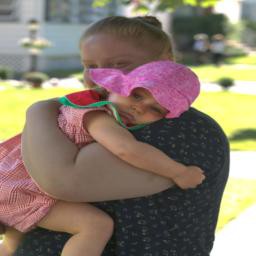}
         \caption{Input}
    \end{subfigure}
    \hfill
    \begin{subfigure}[t]{0.242\columnwidth}
         \centering
         \includegraphics[width=\columnwidth]{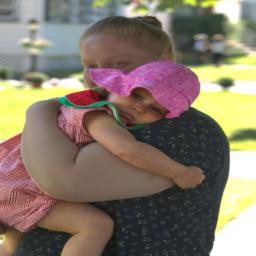}
         \caption{$I_{fore}$}
    \end{subfigure}
    \hfill
    \begin{subfigure}[t]{0.242\columnwidth}
         \centering
         \includegraphics[width=\columnwidth]{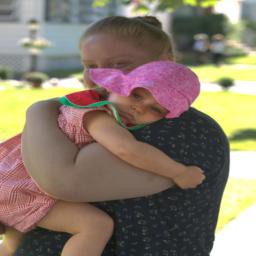}
         \caption{$I_{com}$}
    \end{subfigure}
    \begin{subfigure}[t]{0.242\columnwidth}
        \centering
        \includegraphics[width=\columnwidth]{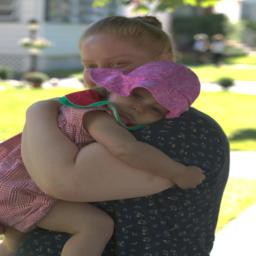}
        \caption{Ours}
    \end{subfigure}
    \vspace{-1em}
    \caption{The influence of different encoder designs.}
    \vspace{-1em}
    \label{fig:encoder}
\end{figure}

\textbf{Loss Function.}
 As shown in Table~\ref{table:ablation} Model ${A}$ to ${C}$, we compare the performance using different loss functions. Since background and foreground domains are different, restricting the loss function on the masked region by using relative $L_1$~($rL_1$) rather than $L_1$ loss helps a lot. Besides, ${L}_{adv}$ are used to improve the realism of the predicted result.  
 
\textbf{Encoder Design $\Phi$.} 
Extracting and learning the global foreground and background features individually~(Ours in Model ${C}$ in Table~\ref{table:ablation}) are also the keys to facilitate the performance of the whole framework. As shown in Table~\ref{table:ablation} and Figure~\ref{fig:encoder}, compared with other alternatives that extract the global features using the foreground region only~($I_{fore}$ in  Model ${D}$) and the full image~($I_{com}$ in Model ${E}$), spatial-separated encoder shows a much better performance due to domain separation.

\begin{figure}[t!]
    \centering
    \begin{subfigure}[b]{0.325\columnwidth}
        \centering
        \includegraphics[width=\columnwidth,height=0.75\columnwidth]{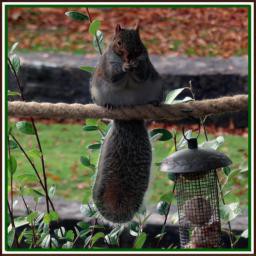}
    \end{subfigure}
    \hfill
    \begin{subfigure}[b]{0.325\columnwidth}
        \centering
        \includegraphics[width=\columnwidth,height=0.75\columnwidth]{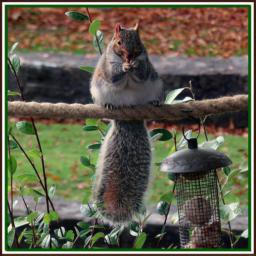}
    \end{subfigure}
    \hfill
    \begin{subfigure}[b]{0.325\columnwidth}
        \centering
        \includegraphics[width=\columnwidth,height=0.75\columnwidth]{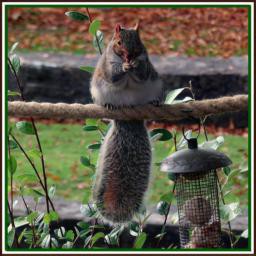}
    \end{subfigure}

    \par\bigskip\vspace*{-1em}
    \begin{subfigure}[b]{0.325\columnwidth}
        \centering
        \includegraphics[width=\columnwidth,height=0.75\columnwidth]{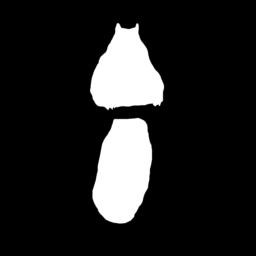}
        \caption{Input\&Mask}
    \end{subfigure}
    \hfill
    \begin{subfigure}[b]{0.325\columnwidth}
        \centering
        \includegraphics[width=\columnwidth,height=0.75\columnwidth]{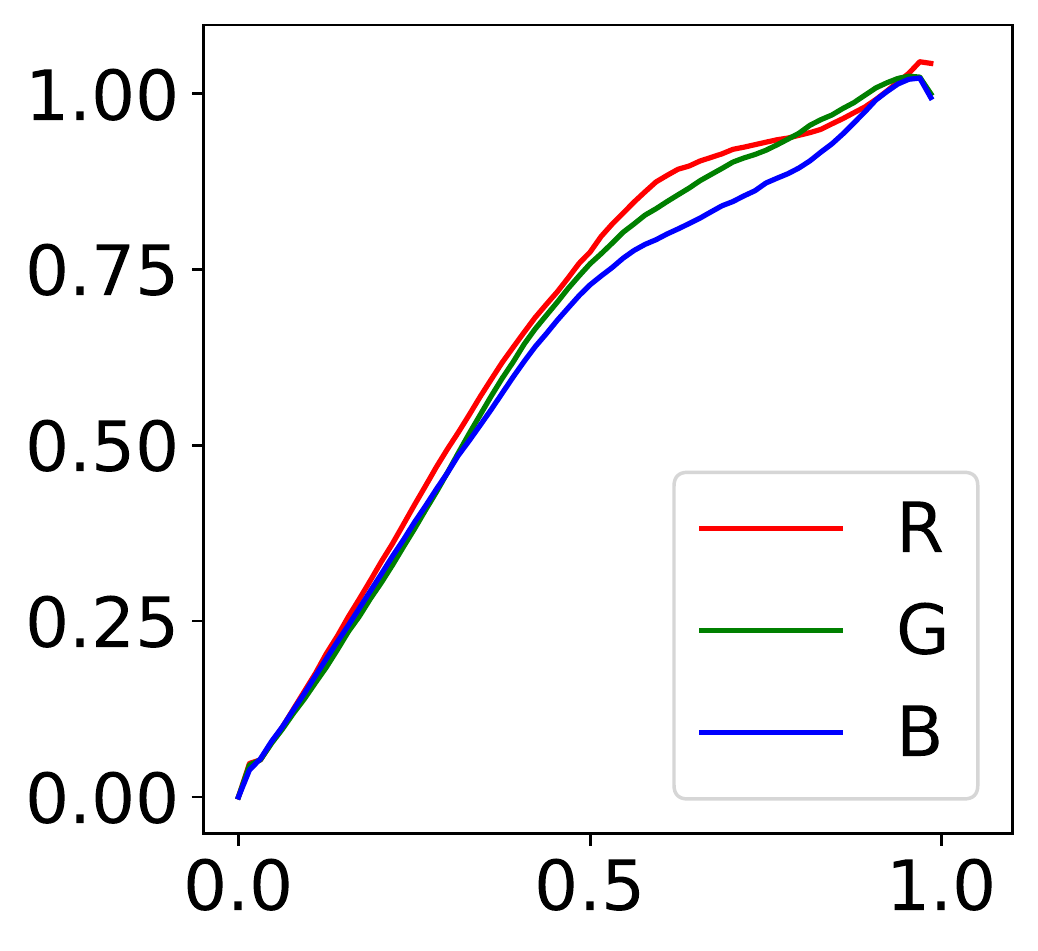}
        \caption{Using \emph{Animal}}
    \end{subfigure}
    \hfill
    \begin{subfigure}[b]{0.325\columnwidth}
        \centering
        \includegraphics[width=\columnwidth,height=0.75\columnwidth]{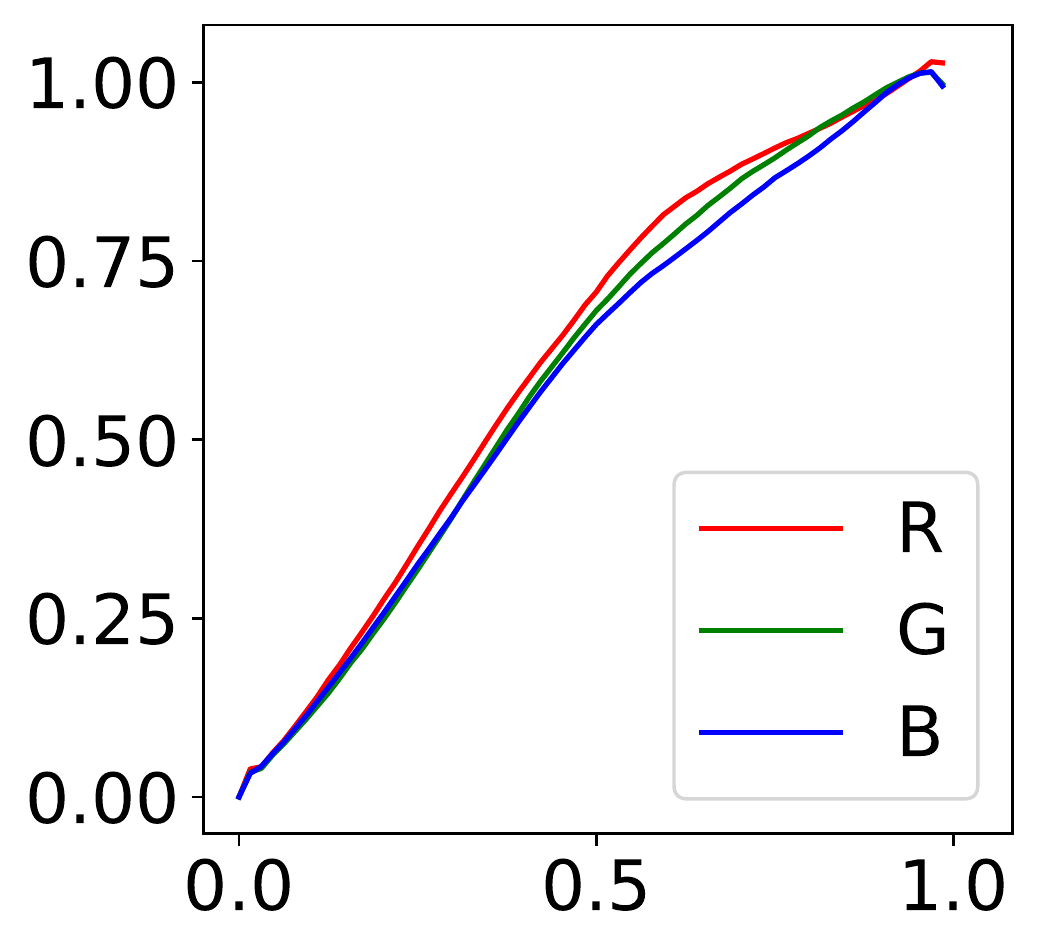}
        \caption{Using \emph{Person}}
    \end{subfigure}
    \caption{Results and rendering curves of SCRM using different foreground semantic labels~(\emph{Person}, \emph{Animal}).}
    \label{fig:seg_labels}
\end{figure}

\begin{figure}[t]
    \centering 
    \begin{subfigure}[b]{0.242\columnwidth}
         \centering
         \includegraphics[width=\columnwidth,height=\columnwidth]{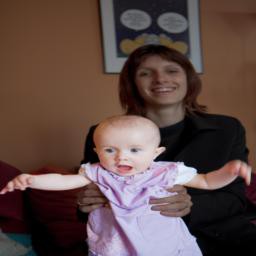}
         \caption{Input}
    \end{subfigure}
    \hfill
    \begin{subfigure}[b]{0.242\columnwidth}
         \centering
         \includegraphics[width=\columnwidth,height=\columnwidth]{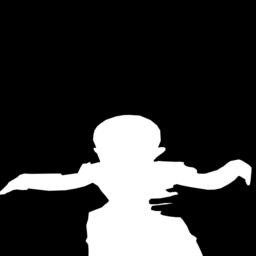}
         \caption{Mask}
    \end{subfigure}
    \hfill
    \begin{subfigure}[b]{0.242\columnwidth}
         \centering
         \includegraphics[width=\columnwidth,height=\columnwidth]{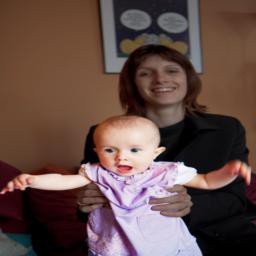}
         \caption{1$^{st}$ Result}
    \end{subfigure}
    \hfill
    \begin{subfigure}[b]{0.242\columnwidth}
         \centering
         \includegraphics[width=\columnwidth,height=\columnwidth]{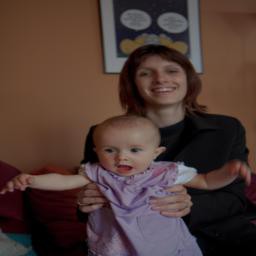}
         \caption{2$^{nd}$ Result}
    \end{subfigure}

    \par\bigskip\vspace*{-1em}
    \begin{subfigure}[t]{0.242\columnwidth}
        \centering
        \includegraphics[width=\columnwidth,height=\columnwidth]{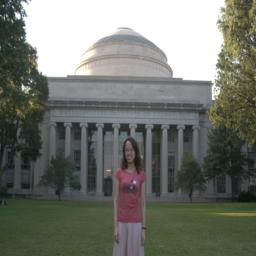}
        \caption{Novel Image}
    \end{subfigure}
    \hfill
    \begin{subfigure}[t]{0.242\columnwidth}
         \centering
         \includegraphics[width=\columnwidth,height=\columnwidth]{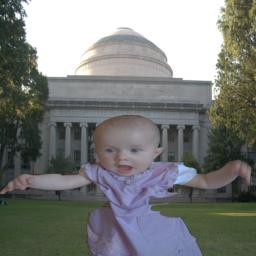}
         \caption{Novel Result}
    \end{subfigure}
    \hfill
    \begin{subfigure}[t]{0.242\columnwidth}
         \centering
         \includegraphics[width=\columnwidth,height=\columnwidth]{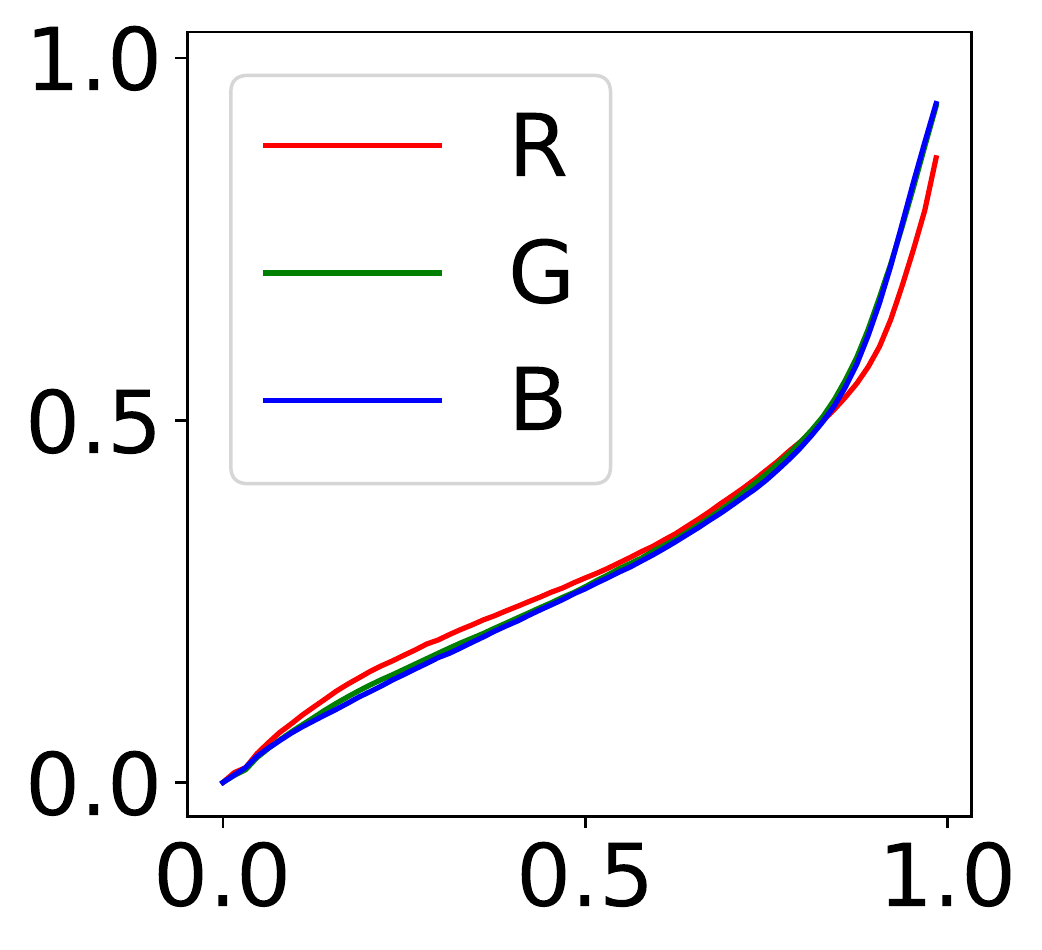}
         \caption{1$^{st}$ Curve}
    \end{subfigure}
    \hfill
    \begin{subfigure}[t]{0.242\columnwidth}
         \centering
         \includegraphics[width=\columnwidth,height=\columnwidth]{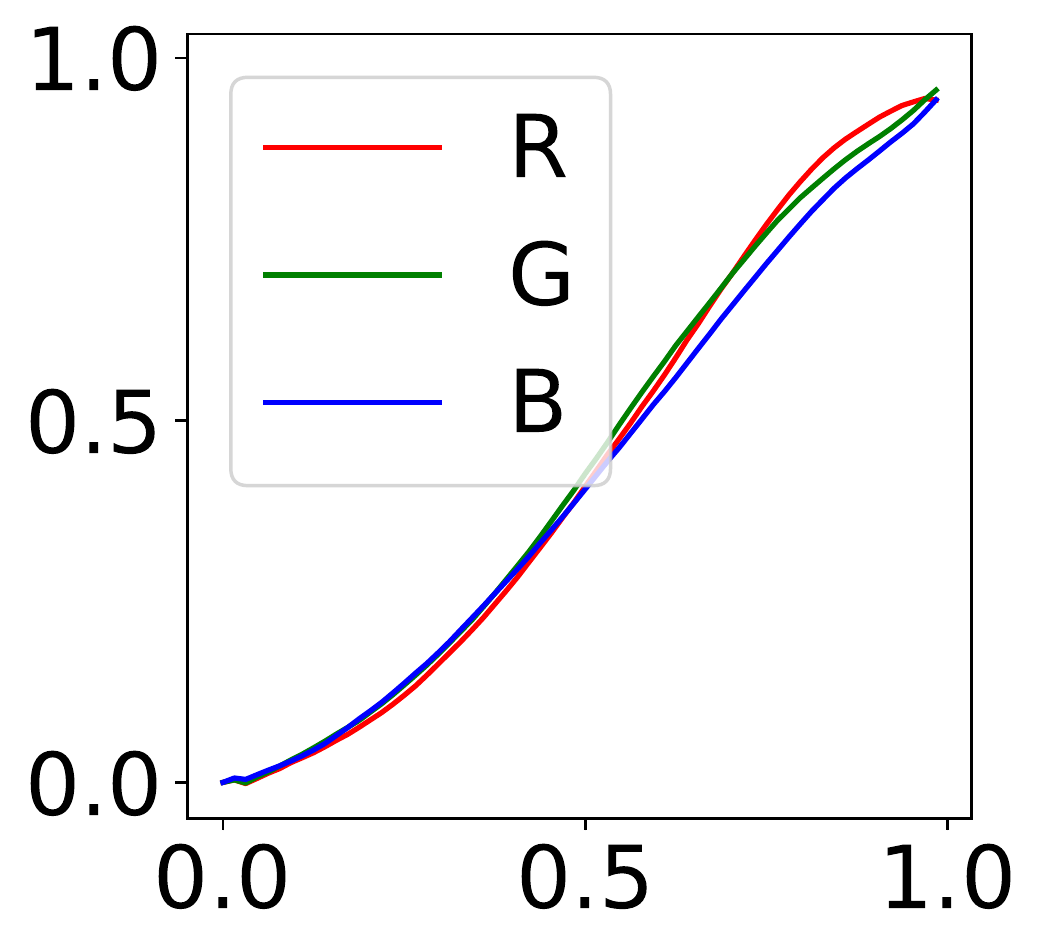}
         \caption{2 $^{nd}$ Curve}
    \end{subfigure}
    \caption{Given a composite image~(a) and its mask~(b), Cascaded-CRM learns to generate different harmonization results~(c, d) via curves~(g, h). Also,our method can harmonize the current foreground via novel backgrounds~(e, f).}
    \label{fig:curve_analysis}
\end{figure}

 \textbf{CRMs.}
 The numerical metrics of different CRMs have been listed in Table~\ref{table:ablation}, both \emph{Cascaded}-CRM (Model~${F}$) and \emph{Cascaded}-SCRM~(Model~${G}$) hugely improve the base model~(Model~${E}$). To further explore the influence of each variants, firstly, we show the importance of semantic labels. As shown in Figure~\ref{fig:seg_labels}, different semantic labels will produce different style curves under the same input. Then, for cascaded refinement, in Figure~\ref{fig:curve_analysis}, cascaded refinement will produce different curves and achieve gradually better performance. Finally, the global color curves enable the proposed method to harmonize images with domain-aware features from novel images. In Figure~\ref{fig:curve_analysis}, a novel background can also guide the harmonization of foreground regions.

\textbf{Comparison with Other Similar Global Editing Methods.}
To further demonstrate the effectiveness of the proposed CRM, we replace our piece-wise linear curve function in CRM by other similar global editing methods in image enhancement~(3DLUT~\cite{3dlut}) and low-light enhancement~(Zero-DCE~\cite{Guo_2021_CVPR}), and compare the performance on HCOCO and iHarmony dataset. As summarized in Table \ref{table:compare_curves}, the piece-wise curve~(Ours) achieves superior performance at all criteria metrics compared to other alternatives. 

\begin{table}[t]
\centering
\resizebox{\columnwidth}{!}{%
\begin{tabular}{@{}l|cccccc@{}}
\toprule
  & \multicolumn{3}{c|}{HCOCO}                  & \multicolumn{3}{c}{iHarmony} \\ \hline
\multicolumn{1}{l|}{Methods} & MSE $\downarrow$   & PSNR $\uparrow$ & \multicolumn{1}{c|}{SSIM$\uparrow$}  & MSE$\downarrow$      & PSNR $\uparrow$   & SSIM$\uparrow$    \\ \hline
\multicolumn{1}{l|}{Zero-DCE~\cite{guo2020zero}}          & 37.22 & 36.64 & \multicolumn{1}{c|}{99.10}  & 75.31    & 34.39   & 98.27   \\
\multicolumn{1}{l|}{3DLUT~\cite{3dlut}}            & 33.22 & 36.95 & \multicolumn{1}{c|}{99.18} & 53.05    & 35.49   & 98.77   \\
\multicolumn{1}{l|}{Ours}                                    & \textbf{29.45}  & \textbf{37.51} & \multicolumn{1}{c|}{\textbf{99.26}}                      & \textbf{45.17}     & \textbf{36.27}   & \textbf{98.87}   \\ 
\bottomrule
\end{tabular}
}
\caption{Quantitative comparison in employing similar global editing methods in our CRM.}
\label{table:compare_curves}
\end{table}

\textbf{The Levels of Curve $L$.}
In the proposed CRM, we approximate the editing curve by a $L$-levels piece-wise linear function. Here, we conduct the ablation experiments to investigate the influence of $L$ by setting $L=\{32, 64, 96, 128\}$ in our S$^2$CRNet model. From Table~\ref{table:ablation_L}, it can be inferred that approximating the curve with more levels improves the harmonizing performance. However, when $L$ is larger than 64, increasing $L$ has minor improvements on HCOCO and even downgrades the performance on the iHarmony dataset. 
It reveals that harmonizing images by a larger $L$ will make the network hard to learn the meaningful color distribution and increase the computational cost. Hence, we set $L=64$ in all models for a trade-off between model performance and memory computation. 

\begin{table}[tbh!]
\centering

\resizebox{\columnwidth}{!}{%
\begin{tabular}{@{}c|ccc|ccc@{}}
\toprule
Dataset    & \multicolumn{3}{c|}{HCOCO} & \multicolumn{3}{c}{iHarmony} \\ \hline
Numer of $L$ & PSNR$\uparrow$    & SSIM$\uparrow$    & MSE$\downarrow$    & PSNR$\uparrow$     & SSIM$\uparrow$    & MSE$\downarrow$     \\ \hline
32         & 37.60    & 99.24   & 29.13  & 36.17    & 98.83   & 48.31   \\
64         & \textbf{37.72}   & \textbf{99.26}   & \textbf{27.40}   & \textbf{36.45}    & \textbf{98.92}   & \textbf{43.20}    \\
96         &  \textbf{37.72}       & \textbf{99.26}        &  27.81      & 36.24    & 98.88   & 46.63    \\
128        & 37.68   & \textbf{99.26}   & 28.58  &   36.19       &  98.86       &  47.00       \\ \bottomrule
\end{tabular}}
\caption{Ablation studies of the level of curve $L$. }
\label{table:ablation_L}
\end{table}

%% file: src/tex/conclusion.tex
\subsection{Discussion and Real-World Application}
\vspace{-.5em}
The proposed framework start a new direction for image harmonization which is efficient, flexible and transparent. As for efficiency, both performance and speed are better than previous methods. With respect to flexibility, images at any resolution can be edited without additional processing, like guided filter~\cite{DIH_Tsai, he2012guided}. As for transparency, our method is a ``white-box" algorithm because the learned curves can be further edited by the user to increase/decrease the harmony.

\subsection{Limitation}
\vspace{-0.5em}
Our method still suffer from some limitations. The global editing curve do not contains any semantic and local information, which restricts further improvement. Since we train our method using limited synthesized dataset but the real-world scenarios are complicated, the performance may be degraded on real world dataset as we have shown some unsatisfactory results in the supplemental materials.

\noindent\textbf{Potential negative impact.} Image harmonization is used to create realistic spliced images, which may be used to create fake media in news or fake data in published papers. However, this task do not change the identity of person or content that could be identified by forgery detection methods~\cite{zhou2018learning}.

\section{Conclusion}
\vspace{-0.5em}
\label{sec:conclusion}
In this paper, we consider image harmonization as a global curve learning problem instead of the pixel-wise prediction for the first time. To this end, we present Spatial-separated Curve Rendering Network~(S$^2$CRNet), a novel framework for efficient and high-resolution image harmonization. In detail, we utilize an efficient backbone to obtain spatial domain-aware features and the extracted features are used to generate the parameters of piece-wise curve function in the proposed curve render model and its variants. Finally, the learnt parameters are used to render the original high-resolution composite foreground. Experiments show the advantages of the proposed framework in terms of efficiency, accuracy and speed.

%% file: supp_arxiv.tex
\appendix

\newcommand{\xiaodong}[1]{{\color{blue}{[xiaodong: #1]}}}
\newcommand\TODO{{\color{red}TODO}}
\newcommand\wfive{0.19\textwidth}
\newcommand\wsix{0.16\textwidth}
\newcommand\wsev{0.135\textwidth}

\newcommand\wweight{0.115\textwidth}

\section{More Implementation Details}

\subsection{The Details of Semantic Labels Extraction}
In the proposed semantic curve rendering module~(SCRM), correct foreground semantic label benefits the performance of image harmonization. However, iHarmony4~\cite{dovenet} do not contain the ground truth labels of the foreground. To obtain the semantic labels, firstly, we get the categories in HCOCO sub-dataset via COCO API~\cite{mscoco}. For the rest sub-datasets, we leverage a semantic segmentation model in \cite{zhou2018semantic} to segment the composite images. Then, we choose the segmented region which has maximal intersection with the foreground mask, and consider it as the category label. Finally, we summarize the distributions of the foreground labels of the whole dataset in Table~\ref{table:domains}. Particularly, we roughly divide the foreground regions into 5 categories, including $\mathit{Person}, \mathit{Vehicle}, \mathit{Animal}, \mathit{Food}$ and others. We argue that this setting is also suitable for the daily usages.

\begin{table}[tbh!]
\centering
\resizebox{\columnwidth}{!}{%
\begin{tabular}{@{}l|c|c|c|c|c@{}}
 \toprule
Classes & HCOCO & HAdobe5k & HFlickr & Hday2night & iHarmony4 \\ \hline
Person     & 13416 & 7274     & 1629    & 0          & 22319     \\
Vehicle    & 4434  & 1338     & 808     & 10         & 6590      \\
Animal     & 7274  & 747      & 675     & 0          & 8696      \\
Food       & 6752  & 280      & 721     & 0          & 7753      \\
Others     & 10952 & 11958    & 4444    & 434        & 27788     \\\hline

Total      & 42828 & 21597    & 8277    & 444        & 73146     \\ \bottomrule
\end{tabular}
}
\vspace{-1em}
\caption{Predicted foreground distributions in iHarmony4.}
\vspace{-1em}
\label{table:domains}
\end{table}

\subsection{The Details of User Study on DIH99.}

\begin{figure}[h]
    \centering
    \includegraphics[width=\columnwidth]{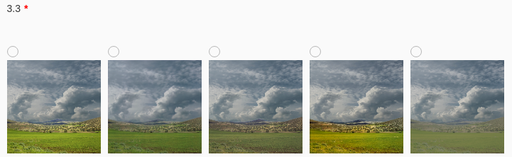}
    \vspace{-2em}
    \caption{The layout of a single image group example in our user study. The displaying order of the composite input and the harmonization results is randomly shuffled without annotations.}
    \vspace{-1em}
    \label{fig:user_study}
\end{figure}

As is discussed in the main paper, to evaluate the effectiveness on real-world scenarios, we conduct subjective user study to compare our proposed method with baseline methods~(DIH~\cite{DIH_Tsai}, DoveNet~\cite{dovenet} and BargainNet~\cite{bargainet}) on  the DIH99 real composite dataset. In detail, we invite 18 participants with different ages and genders for subjective experiments. As shown in Figure~\ref{fig:user_study}, each participant can see a set of image groups and each group includes the original composite input and the harmonized results that generated by DIH, DoveNet, BargainNet and the proposed S$^2$CRNet. Then, we let them to select the most favorable result among different images in each image group, contributing 18$\times$99 groups result in total. The results have been listed in Table~2 of the main paper.

\section{More Experiments}

\newcommand{\imnamehr}[2]{src/figures_supp/HR/#1/#2.jpg}
\newcommand{\methodshr}{Input,DoveNet,BargainNet,S2AM,Ours,Ours_vgg,target}
\newcommand{\imgshr}{2048,1024,512}

\begin{figure*}[t]
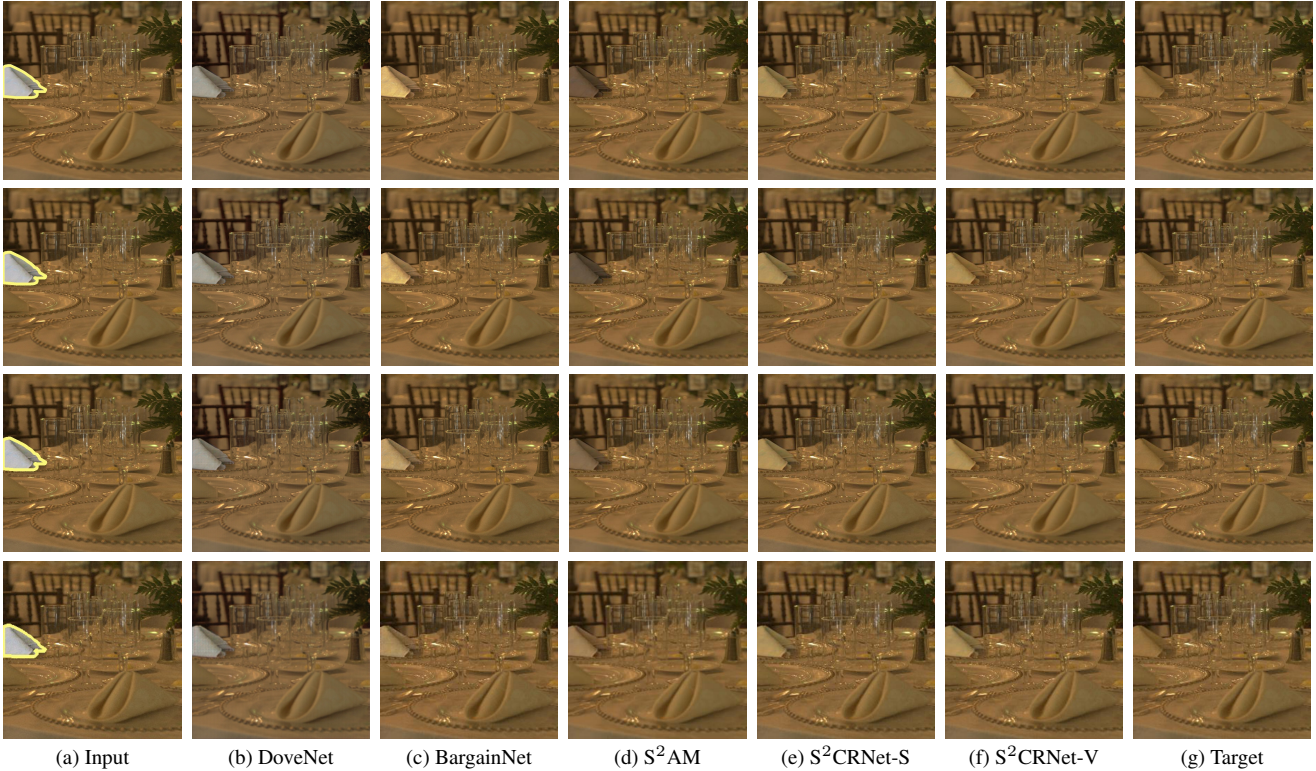

\centering     %
\makeatletter
\@for\img:=\imgshr\do{
    \@for\method:=\methodshr\do{
    \begin{subfigure}[t]{0.135\textwidth}
            \centering
     \includegraphics[width=\columnwidth,height=\columnwidth]{\imnamehr{\method}{\img}}
    \end{subfigure}\hfill
    }
    \vfill\vspace{0.25em}
}
\makeatother

\begin{subfigure}[t]{0.135\textwidth}\centering
     \includegraphics[width=\columnwidth,height=\columnwidth]{\imnamehr{Input}{256}}\caption{Input}
\end{subfigure}\hfill
\begin{subfigure}[t]{0.135\textwidth}\centering
     \includegraphics[width=\columnwidth,height=\columnwidth]{\imnamehr{DoveNet}{256}}\caption{DoveNet}
\end{subfigure}\hfill
\begin{subfigure}[t]{0.135\textwidth}\centering
     \includegraphics[width=\columnwidth,height=\columnwidth]{\imnamehr{BargainNet}{256}}\caption{BargainNet}
\end{subfigure}\hfill
\begin{subfigure}[t]{0.135\textwidth}\centering
     \includegraphics[width=\columnwidth,height=\columnwidth]{\imnamehr{S2AM}{256}}\caption{S$^2$AM}
\end{subfigure}\hfill
\begin{subfigure}[t]{0.135\textwidth}\centering
     \includegraphics[width=\columnwidth,height=\columnwidth]{\imnamehr{Ours}{256}}\caption{S$^2$CRNet-S}
\end{subfigure}\hfill
\begin{subfigure}[t]{0.135\textwidth}\centering
     \includegraphics[width=\columnwidth,height=\columnwidth]{\imnamehr{Ours}{256}}\caption{S$^2$CRNet-V}
\end{subfigure}\hfill
\begin{subfigure}[t]{0.135\textwidth}\centering
     \includegraphics[width=\columnwidth,height=\columnwidth]{\imnamehr{target}{256}}\caption{Target}\hfill\vfill
\end{subfigure}
\vspace{-1em}
\caption{Qualitative comparisons with existing methods in harmonizing images at different resolutions. Here, we resize the all the images to the same resolutions for presentation. The original input images are $256\times256$, $512\times512$, $1024\times1024$, $2048\times2048$ from the bottom up successively. We mark the composite foreground mask as yellow region. S$^2$CRNet-S and S$^2$CRNet-V denote our method employing SqueezeNet and VGG16 backbone, respectively.}
\label{fig:supp_hr}
\end{figure*}

\subsection{The Effectiveness of Different Backbone.}
Stronger backbone enables the networks to learn better. We evaluate the performance of different backbones in our framework as shown in Table~\ref{table:backbones}. We find that more complicated structures such as VGG16~\cite{simonyan2014very} perform much better than the smaller backbones, but it lacks efficiency as reported in the main paper. Also, complicated structures need more epochs to be convergent~(for example, SqueezeNet-based method only needs 20 epoch to get the best result while VGG16 achieves the best performance at 48 epoch.). Thus, we report the results of best performance~(VGG16 backbone) and the most efficient model~(SqueezeNet backbone) in the main paper.

\begin{table}[tbh!]
\resizebox{\columnwidth}{!}{%
\begin{tabular}{ll|lll}
\toprule
\multicolumn{1}{l|}{Backbones} & Param  & PSNR $\uparrow$ & SSIM $\uparrow$ & MSE $\downarrow$   \\ \hline
\multicolumn{1}{l|}{SqueezeNet~\cite{iandola2016squeezenet}}        & 0.95M  & 36.45 & 98.92 & 43.20  \\
\multicolumn{1}{l|}{AlexNet~\cite{alexnet}}           & 2.79M  & 35.82 & 98.80  & 50.79 \\
\multicolumn{1}{l|}{ResNet18~\cite{resnet}}          & 11.8M  & 36.55 & 98.92 & 41.04 \\
\multicolumn{1}{l|}{VGG16~\cite{simonyan2014very}}          & 15.14M & 37.18 & 99.01 & 35.58 \\ \bottomrule
\end{tabular}}
\vspace{-1em}
\caption{Performance of different backbones in the proposed S$^2$CRNet. All experiments are trained and evaluated on iHarmony dataset under the same configurations.}
\label{table:backbones}
\end{table}

\subsection{Visual Comparison on High-Resolution Images}
Our method shows the resolution-invariant results that benefits from the proposed curve-based  framework. Here, we visualize an example to show the influence of the input resolution in different methods. Similar to the high-resolution image harmonization experiments in the primary paper, we compare our method with other baseline methods~\cite{dovenet, bargainet, s2am} in harmonizing images at different resolutions including the square of 256, 512, 1024 and 2048.  As shown in Figure~\ref{fig:supp_hr}, due to the changes of reception fields, the other state-of-the-art methods show unstable results. Differently, both the proposed S$^2$CRNet-SqueezeNet and S$^2$CRNet-VGG16 get more stable and favorable results, while the others show downgraded harmonization qualities as the resolutions increase.

\subsection{Harmonization Performance on CPU.}
Our method also shows good speed on CPU devices, which enables our method to run on the device side without any cloud computation. To this end, we compare the proposed S$^2$CRNet with other baseline methods~\cite{s2am,dovenet,bargainet,Guo_2021_CVPR} in harmonizing different resolution images using the same experimental environment~(Intel i7-10700K CPU with 16 GB RAM on Ubuntu 18.04). Here, we choose the default SqueezeNet backbone in the proposed S$^2$CRNet for efficiency. The evaluations are conducted on the 50 images in HAdobe5k sub-dataset~\cite{dovenet} and we present the average processing time in Table~\ref{table:cpu}. The quantitative results show that our method achieves the fastest performance when operating on the CPU, and also outperforms other baselines by a large margin as the image resolution increases. Notice that our method also shows better performance than these methods as discussed in the main paper.

\begin{table}[t]
\resizebox{\columnwidth}{!}{%
\begin{tabular}{@{}c|ccccc@{}}
\toprule
Resolution & S$^2$AM  & DoveNet & BargainNet & IIH   & S$^2$CRNet \\ \hline
256$\times$256        & 0.25s & 0.05s   & 0.21s      & 1.17s & \textbf{0.03s}     \\
512$\times$512        & 0.85s & 0.18s   & 0.75s      & 8.02s & \textbf{0.06s}     \\
1024$\times$1024       & 3.93s & 0.79s   & 3.23s      & NA    & \textbf{0.47s}     \\
2048$\times$2048       & NA    & 3.19s   & 13.06s     & NA    & \textbf{2.60s}     \\ \bottomrule
\end{tabular}
}
\caption{Average processing time on the \textbf{CPU} under different image resolution. The best results are marked as boldface and the ``NA" denotes running out of memory in our experiment.}
\label{table:cpu}
\end{table}

\newcommand{\imnamefsu}[2]{src/figures_supp/failed_cases/#1/test#2_Su}
\newcommand{\imnamef}[2]{src/figures_supp/failed_cases/#1/test_#2}
\newcommand{\methodsf}{Inputs,DIH,DoveNet,BargainNet,Ours,Ours_vgg}

\newcommand{\imgsfSu}{48}
\newcommand{\imgsf}{18,35}

\begin{figure*}[t]
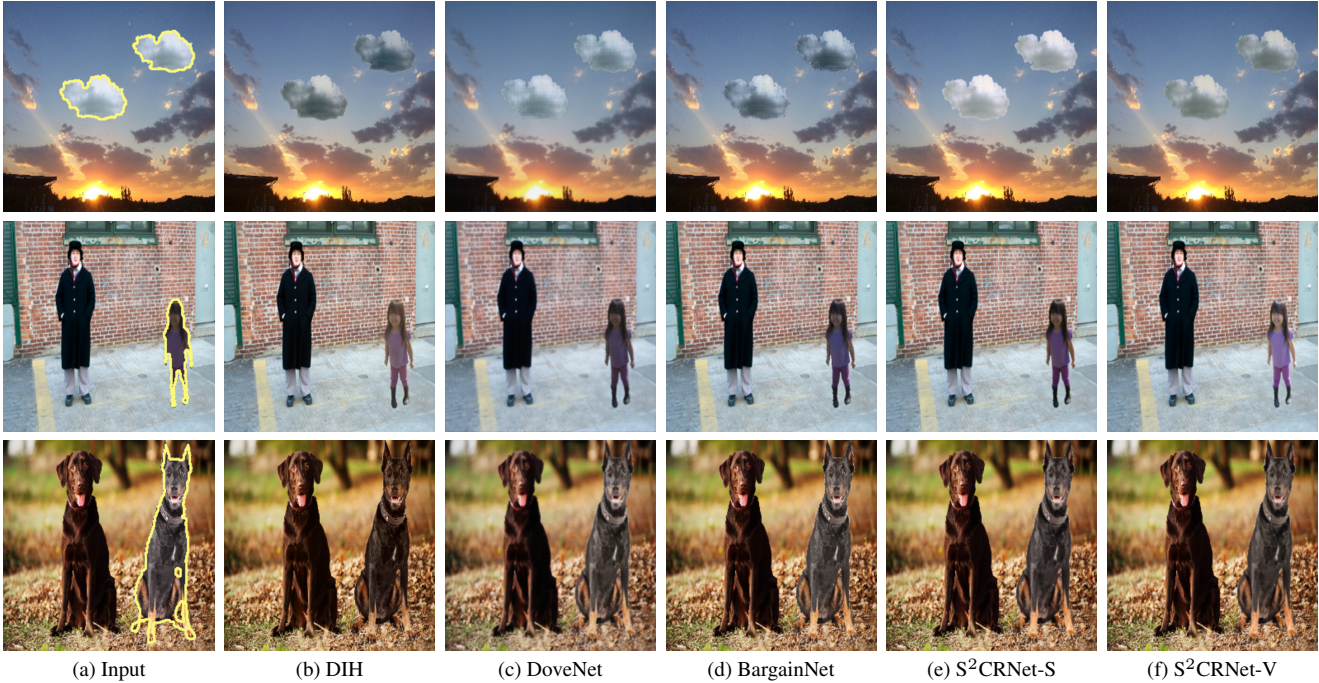

\centering     %
\makeatletter

\@for\img:=\imgsf\do{
    \@for\method:=\methodsf\do{
    \begin{subfigure}[b]{\wsix}
        \centering
           \includegraphics[width=\columnwidth,height=\columnwidth]{\imnamef{\method}{\img}}
    \end{subfigure}\hfill
    }\vfill\vspace{0.25em}
}

\begin{subfigure}[t]{\wsix}\centering
    \includegraphics[width=\columnwidth,height=\columnwidth]{\imnamef{Inputs}{45}} \caption{Input}
\end{subfigure}\hfill
\begin{subfigure}[t]{\wsix}\centering
     \includegraphics[width=\columnwidth,height=\columnwidth]{\imnamef{DIH}{45}}\caption{DIH}
\end{subfigure}\hfill
\begin{subfigure}[t]{\wsix}\centering
     \includegraphics[width=\columnwidth,height=\columnwidth]{\imnamef{DoveNet}{45}}\caption{DoveNet}
\end{subfigure}\hfill
\begin{subfigure}[t]{\wsix}\centering
     \includegraphics[width=\columnwidth,height=\columnwidth]{\imnamef{BargainNet}{45}}\caption{BargainNet}
\end{subfigure}\hfill
\begin{subfigure}[t]{\wsix}\centering
     \includegraphics[width=\columnwidth,height=\columnwidth]{\imnamef{Ours}{45}}\caption{S$^2$CRNet-S}
\end{subfigure}\hfill
\begin{subfigure}[t]{\wsix}\centering
     \includegraphics[width=\columnwidth,height=\columnwidth]{\imnamef{Ours_vgg}{45}}\caption{S$^2$CRNet-V}
\end{subfigure}\hfill\vfill
\makeatother
\caption{Failure harmonization cases of the proposed S$^2$CRNet and  baseline methods~\cite{DIH_Tsai, dovenet, bargainet} on DIH99 dataset. We mark the composite foreground mask as yellow region. S$^2$CRNet-S and  S$^2$CRNet-V  denote our method employing SqueezeNet and VGG16 backbone, respectively.}
\label{fig:failure}
\end{figure*}

\subsection{Failure Harmonization Cases}
As our models are trained and validated on the synthetic iHarmony4 dataset~\cite{dovenet} following previous works~\cite{dovenet,bargainet}, it might encounter some unsatisfactory harmonization results when applying our models in the complicated real scenarios. Here, we present some failure results in harmonizing the real composite samples in DIH99~\cite{DIH_Tsai}. As shown in Figure~\ref{fig:failure}, the inserted foregrounds are not perfectly compatible with the corresponding background images. For example, in the first and the last example, our methods create worse results than DIH. It might be because DIH has a strong ability for scene understanding as they learn harmonization and segmentation jointly. Thus, the inserted cloud and dogs in the examples can match the style of the similar class in the background while in contrast, our methods lack the pixel-wise semantic relationship of both foreground and background. Considering the second example, although our method~(S$^2$CRNet-V) produces better results, it still looks unreal and can be easily distinguished by the human eyes. It might be improved by adding the soft shadow to the image. For this case, we argue that the community needs to define a new scheme for image harmonization beyond the current settings.

\begin{table*}[tbh!]
\centering
\begin{tabular}{@{}c|l|cc|cc|cc|cc@{}}
\toprule
\multicolumn{2}{c|}{Foreground Ratios} & \multicolumn{2}{c|}{0\%-5\%} & \multicolumn{2}{c|}{5\%-15\%} & \multicolumn{2}{c|}{15\%-100\%} & \multicolumn{2}{c}{0\%-100\%} \\ \hline
\multicolumn{2}{c|}{Evaluation metric} & MSE $\downarrow$         & fMSE$\downarrow$          & MSE$\downarrow$           & fMSE$\downarrow$          & MSE$\downarrow$            & fMSE$\downarrow$           & MSE$\downarrow$           & fMSE$\downarrow$          \\ \hline
\multicolumn{2}{l|}{Input composition}   & 28.51        & 1208.86       & 119.19        & 1323.23       & 577.58         & 1887.05        & 172.47        & 1387.30        \\
\multicolumn{2}{l|}{Xue~\etal~\cite{intro_illumination}} & 41.52        & 1481.59       & 120.62        & 1309.79       & 444.65         & 1467.98        & 150.53        & 1433.21       \\
\multicolumn{2}{l|}{Lalonde \& Efros~\cite{intro_color1}}          & 31.24        & 1325.96       & 132.12        & 1459.28       & 479.53         & 1555.69        & 155.87        & 1411.40        \\
\multicolumn{2}{l|}{Zhu~\etal~\cite{zhu2015learning}}        & 33.30         & 1297.65       & 145.14        & 1577.70        & 682.69         & 2251.76        & 204.77        & 1580.17       \\
\multicolumn{2}{l|}{DIH\cite{DIH_Tsai}}               & 18.92        & 799.17        & 64.23         & 725.86        & 228.86         & 768.89         & 76.77         & 773.18        \\
\multicolumn{2}{l|}{DoveNet\cite{dovenet}}           & 14.03        & 591.88        & 44.90          & 504.42        & 152.07         & 505.82         & 52.36         & 549.96        \\
\multicolumn{2}{l|}{S$^2$AM\cite{s2am}}              & 13.51        & 509.41        & 41.79         & 454.21        & 137.12         & 449.81         & 48.00           & 481.79        \\
\multicolumn{2}{l|}{BargainNet\cite{bargainet}}        & 10.55        & 450.33        & 32.13         & 359.49        & \underline{109.23}         & \underline{353.84}         & \underline{37.82}         & 405.23        \\
\multicolumn{2}{l|}{IIH\cite{Guo_2021_CVPR}}        & 9.97        & 441.02        & 31.51        & 363.61        & {110.22}         & {354.84}         & {38.71}         & 400.29        \\
\multicolumn{2}{l|}{RainNet\cite{Ling_2021_CVPR}}           & 11.66        & 550.38        & 32.05         & 378.69        & 117.41         & 389.81          & 40.29         & 469.61         \\ \hline
\multicolumn{2}{l|}{S$^2$CRNet-SqueezeNet}           & \underline{8.42}         & \underline{301.97}        & \underline{29.74}         & \underline{336.24}        & 126.56         & 405.13         & 43.21          & \underline{336.99}        \\ 
\multicolumn{2}{l|}{S$^2$CRNet-VGG16}           & \textbf{6.80}         & \textbf{239.94}        & \textbf{25.37}         & \textbf{271.70}        & \textbf{103.42}       & \textbf{333.96}         & \textbf{35.58}          & \textbf{274.99}        \\ \bottomrule
\end{tabular}
\vspace{-1em}
\caption{Foreground Harmonization Comparisons on iHarmony4. The fMSE measures the mean square error scores of the harmonized foreground regions.  The best and the second best are marked as boldface and underline respectively. }
\label{table:fmse}
\end{table*}

\subsection{The Results on Different Foreground Ratios.}

The differences of the composite foregrounds are also important in our task since the background is totally the same. Thus, we further compare the proposed S$^2$CRNet~(including SqueezeNet~\cite{iandola2016squeezenet} and VGG16~\cite{simonyan2014very} backbones) with other state-of-art image harmonization approaches in different foreground ratio ranges, and the quantitative results on iHarmony4 dataset are summarized in Table~\ref{table:fmse}. Following previous methods~\cite{dovenet,bargainet,Ling_2021_CVPR}, we employ mean square error~(MSE) and foreground mean square error~(fMSE) as evaluation metrics, where fMSE measures the MSE scores of the harmonized foreground regions. We follow previous works~\cite{dovenet,bargainet} to evaluate the performance in four different foreground ranges, including 0\% to 5\%, 5\% to 15\%, 15\% to 100\% and overall results.
As shown in Table \ref{table:fmse}, the performance of all the models will be downgraded as the foreground ratios increase. Nevertheless, our S$^2$CRNet-SqueezeNet achieves the best performance in most of the foreground ratio intervals especially on small foreground regions~(0\%-5\% and 5\%-15\% foreground ratios). Furthermore, when employing VGG16 backbone~(S$^2$CRNet-VGG16), our method achieves state-of-art performance and outperforms other methods by a large margin in all the foreground ratio intervals.

\subsection{More Rendering Curves Visualization}
We present more visual results to visualize the preliminary rendering curves and the curves in cascaded refinements. As shown in Figure \ref{fig:supp_curve}, the curves generated by the CRM are different according to various input samples, which demonstrates that the S$^2$CRNet can produce the practical curve parameters for each images via the deep features. Also, for cascaded refinement, the curves in each stage are also different and these stage-aware curves contribute to the improvement of the harmonization performance according to the visualized results of different stages in Figure \ref{fig:supp_curve}.

\subsection{Visualized Comparison in DIH99}
To demonstrate the effectiveness of the proposed method on real-world scenarios, we further evaluate the proposed S$^2$CRNet with two backbones (SqueezeNet\cite{iandola2016squeezenet} and VGG16~\cite{simonyan2014very}) and the baseline methods~(DIH~\cite{DIH_Tsai}, DoveNet~\cite{dovenet} and BargainNet~\cite{bargainet}) on DIH99 real composite dataset, and visualize the harmonization results in Figure~\ref{fig:supp_real}. As shown in Figure~\ref{fig:supp_real}, the proposed efficient S$^2$CRNet can also achieve favorable results on real composite images compared to other presented methods, showing the reliable generalization in real-scenario applications.

\subsection{More Visual Results on iHarmony4 Dataset}
Given some composite images and their foreground masks, in Figure~\ref{fig:supp_all}, we present more harmonized results generated by methods including S$^2$AM~\cite{s2am}, DoveNet~\cite{dovenet}, BargainNet~\cite{bargainet} and our S$^2$CRNet on iHarmony4 dataset. Compared with the other baselines, both S$^2$CRNet-SqueezeNet and S$^2$CRNet-VGG16 can generate more harmonious results and also maintain visual similarities with the target natural images.

\newcommand{\imnamecurves}[2]{src/figures_supp/more_curves/#1/#2}
\newcommand{\methodscurve}{inputs,curve1,curve1_img,curve2,curve2_img}
\newcommand{\imgscurve}{a1580_1_4,c496166_1092610_1,f213_1_1,a3839_1_2}

\begin{figure*}[tb!]
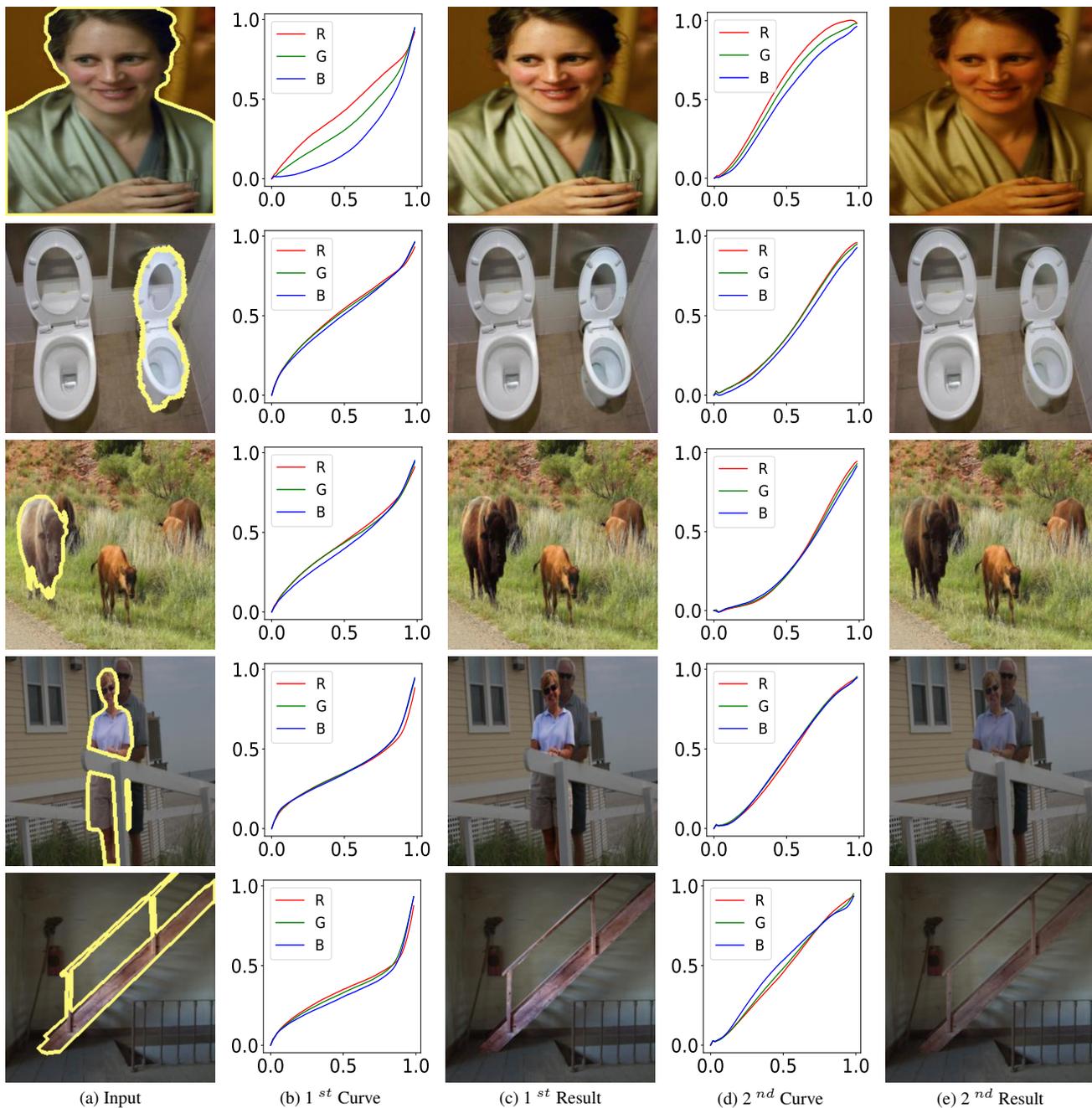

\centering     %
\makeatletter
\@for\img:=\imgscurve\do{
    \@for\method:=\methodscurve\do{
    \begin{subfigure}[t]{\wfive}
    \centering
    \includegraphics[width=\columnwidth,height=\columnwidth]{\imnamecurves{\method}{\img}}
    \end{subfigure}\hfill
    }
    \vfill\vspace{0.25em}
}

\begin{subfigure}[t]{\wfive}\centering
     \includegraphics[width=\columnwidth,height=\columnwidth]{\imnamecurves{inputs}{a3898_1_2}}\caption{Input}
\end{subfigure}\hfill
\begin{subfigure}[t]{\wfive}\centering
     \includegraphics[width=\columnwidth,height=\columnwidth]{\imnamecurves{curve1}{a3898_1_2}}\caption{1 $^{st}$ Curve}
\end{subfigure}\hfill
\begin{subfigure}[t]{\wfive}\centering
     \includegraphics[width=\columnwidth,height=\columnwidth]{\imnamecurves{curve1_img}{a3898_1_2}}\caption{1 $^{st}$ Result}
\end{subfigure}\hfill
\begin{subfigure}[t]{\wfive}\centering
     \includegraphics[width=\columnwidth,height=\columnwidth]{\imnamecurves{curve2}{a3898_1_2}}\caption{2 $^{nd}$ Curve}
\end{subfigure}\hfill
\begin{subfigure}[t]{\wfive}\centering
     \includegraphics[width=\columnwidth,height=\columnwidth]{\imnamecurves{curve2_img}{a3898_1_2}}\caption{2 $^{nd}$ Result}
\end{subfigure}\hfill
\makeatother
\vspace{-1em}
\caption{More visualized results of the cascaded rendering curves generated by S$^2$CRNet. We mark the composite foreground mask as yellow region.}
\label{fig:supp_curve}
\end{figure*}

\newcommand{\imnamein}[2]{src/figures_supp/real_results/#1/test#2_Su}
\newcommand{\imnamec}[2]{src/figures_supp/real_results/#1/test_#2}

\newcommand{\xmethods}{Inputs,DIH,DoveNet,BargainNet,Ours,Ours_vgg}

\newcommand{\imgsSu}{06,03}
\newcommand{\imgsX}{03,04,12,29}

\begin{figure*}[!tb]
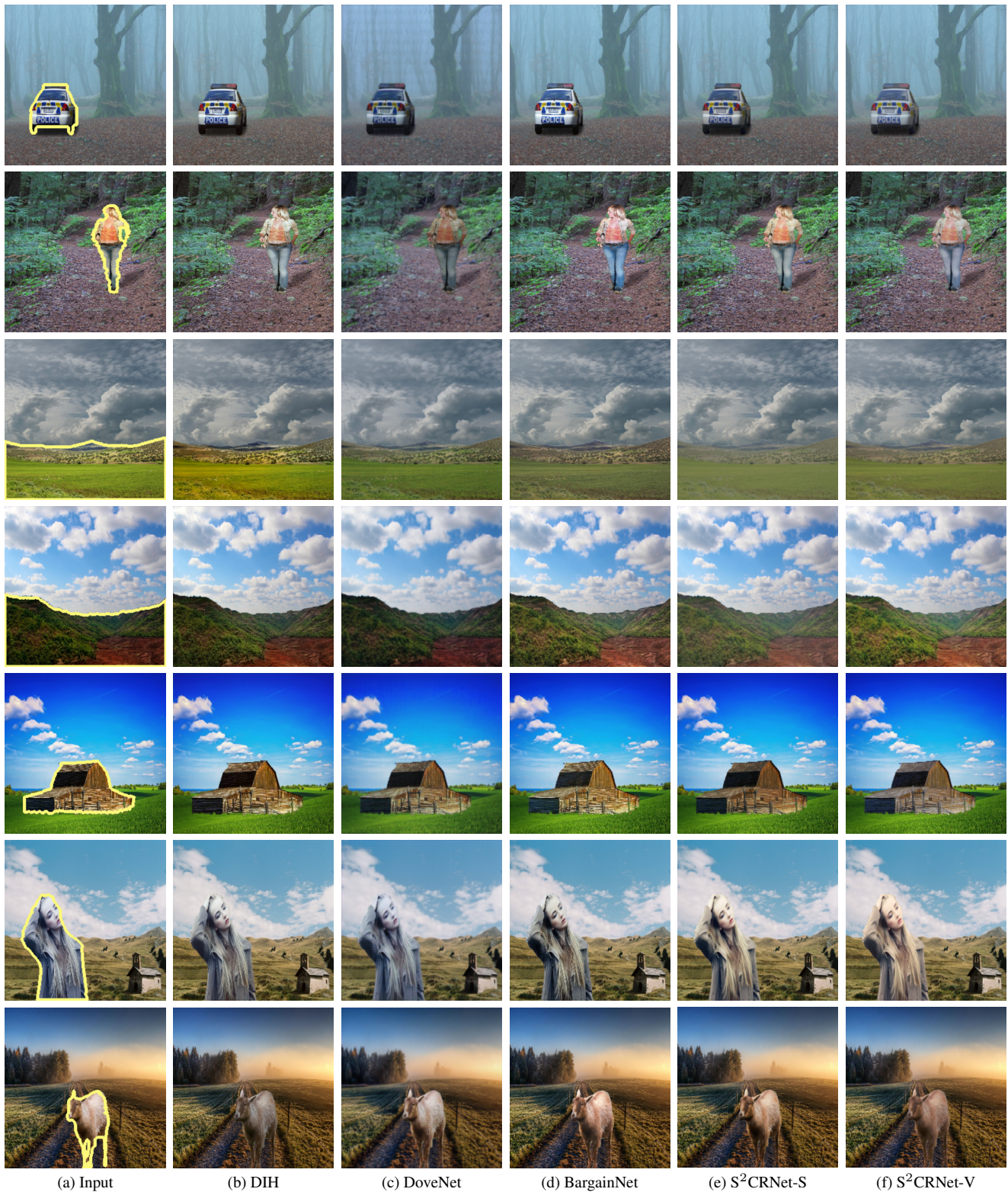

\centering     %
\makeatletter
\@for\img:=\imgsSu\do{
    \@for\method:=\xmethods\do{
    \begin{subfigure}[b]{\wsix}\centering
           \includegraphics[width=\columnwidth,height=\columnwidth]{\imnamein{\method}{\img}}
    \end{subfigure}\hfill
    }\vfill\vspace{0.25em}
}
\@for\img:=\imgsX\do{
    \@for\method:=\xmethods\do{
    \begin{subfigure}[b]{\wsix}
        \centering
           \includegraphics[width=\columnwidth,height=\columnwidth]{\imnamec{\method}{\img}}
    \end{subfigure}\hfill
    }\vfill\vspace{0.25em}
}

\begin{subfigure}[b]{\wsix}\centering
    \includegraphics[width=\columnwidth,height=\columnwidth]{\imnamec{Inputs}{44}} \caption{Input}
\end{subfigure}\hfill
\begin{subfigure}[b]{\wsix}\centering
     \includegraphics[width=\columnwidth,height=\columnwidth]{\imnamec{DIH}{44}}\caption{DIH}
\end{subfigure}\hfill
\begin{subfigure}[b]{\wsix}\centering
     \includegraphics[width=\columnwidth,height=\columnwidth]{\imnamec{DoveNet}{44}}\caption{DoveNet}
\end{subfigure}\hfill
\begin{subfigure}[b]{\wsix}\centering
     \includegraphics[width=\columnwidth,height=\columnwidth]{\imnamec{BargainNet}{44}}\caption{BargainNet}
\end{subfigure}\hfill
\begin{subfigure}[b]{\wsix}\centering
     \includegraphics[width=\columnwidth,height=\columnwidth]{\imnamec{Ours}{44}}\caption{S$^2$CRNet-S}
\end{subfigure}\hfill
\begin{subfigure}[b]{\wsix}\centering
     \includegraphics[width=\columnwidth,height=\columnwidth]{\imnamec{Ours_vgg}{44}}\caption{S$^2$CRNet-V}
\end{subfigure}\hfill\vfill
\makeatother
\caption{More comparisons with baseline methods~\cite{DIH_Tsai, dovenet, bargainet} on DIH99 dataset. We mark the composite foreground mask as yellow region. S$^2$CRNet-S and  S$^2$CRNet-V denote our method employing SqueezeNet and VGG16 backbone, respectively.}
\label{fig:supp_real}
\end{figure*}

\newcommand{\imnameall}[2]{src/figures_supp/compare_all/#1/#2}
\newcommand{\methodsall}{inputs,DoveNet,BargainNet,s2am,Ours,Ours_vgg,Target}
\newcommand{\imgsall}{a3600_1_5,a4438_1_5,c36034_1080575_2,c458603_584367_2,d90000002-24_1_18,f1091_1_2,f3472_1_1}

\begin{figure*}[t]
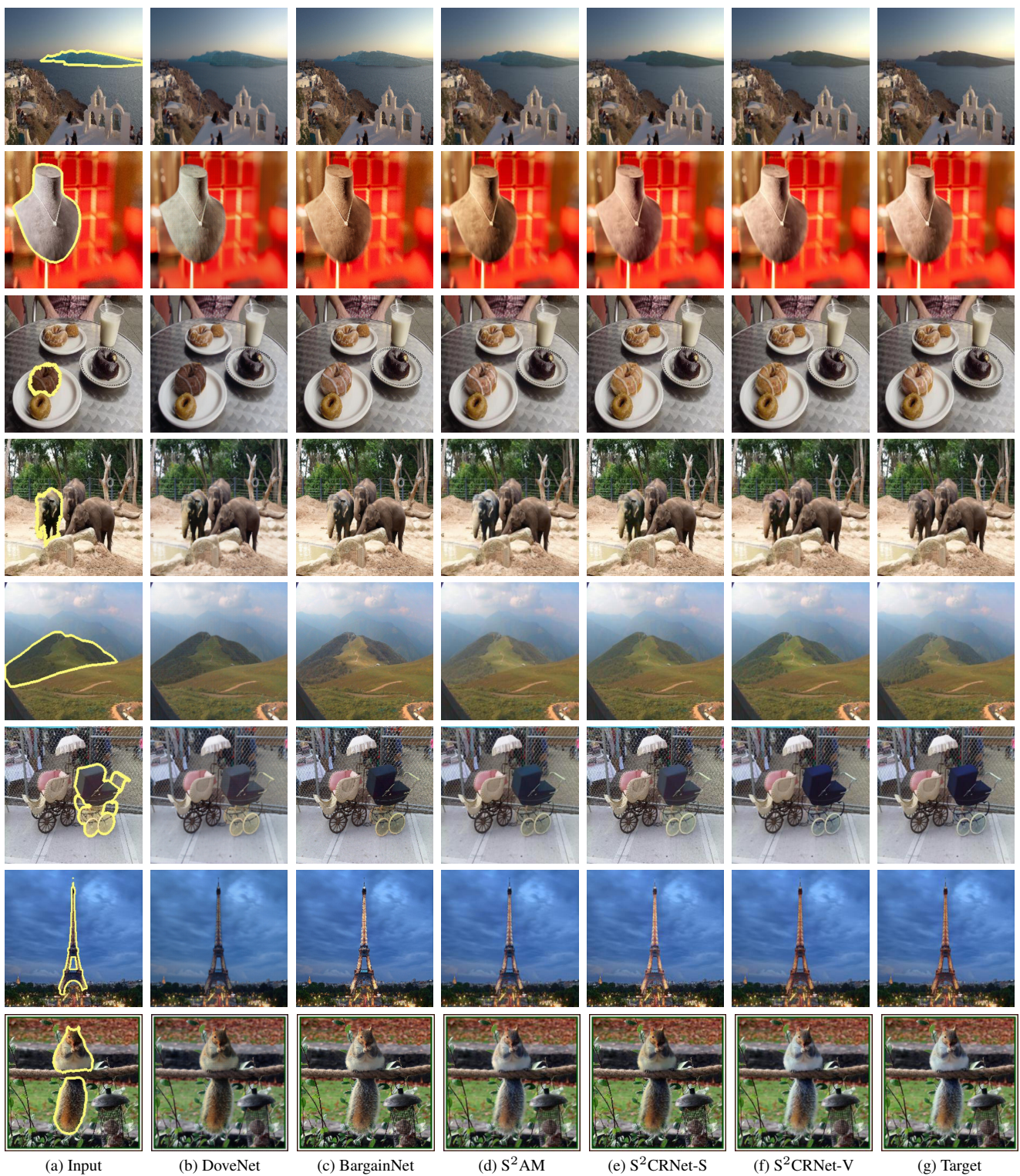

\centering     %
\makeatletter
\@for\img:=\imgsall\do{
    \@for\method:=\methodsall\do{
    \begin{subfigure}[t]{\wsev}
            \centering
     \includegraphics[width=\columnwidth,height=\columnwidth]{\imnameall{\method}{\img}}
    \end{subfigure}\hfill
    }
    \vfill\vspace{0.25em}
}
\makeatother

\begin{subfigure}[t]{\wsev}\centering
     \includegraphics[width=\columnwidth,height=\columnwidth]{\imnameall{inputs}{f1182_1_1}}\caption{Input}
\end{subfigure}\hfill
\begin{subfigure}[t]{\wsev}\centering
     \includegraphics[width=\columnwidth,height=\columnwidth]{\imnameall{DoveNet}{f1182_1_1}}\caption{DoveNet}
\end{subfigure}\hfill
\begin{subfigure}[t]{\wsev}\centering
     \includegraphics[width=\columnwidth,height=\columnwidth]{\imnameall{BargainNet}{f1182_1_1}}\caption{BargainNet}
\end{subfigure}\hfill
\begin{subfigure}[t]{\wsev}\centering
     \includegraphics[width=\columnwidth,height=\columnwidth]{\imnameall{s2am}{f1182_1_1}}\caption{S$^2$AM}
\end{subfigure}\hfill
\begin{subfigure}[t]{\wsev}\centering
     \includegraphics[width=\columnwidth,height=\columnwidth]{\imnameall{Ours}{f1182_1_1}}\caption{S$^2$CRNet-S}
\end{subfigure}\hfill
\begin{subfigure}[t]{\wsev}\centering
     \includegraphics[width=\columnwidth,height=\columnwidth]{\imnameall{Ours_vgg}{f1182_1_1}}\caption{S$^2$CRNet-V}
\end{subfigure}\hfill
\begin{subfigure}[t]{\wsev}\centering
     \includegraphics[width=\columnwidth,height=\columnwidth]{\imnameall{Target}{f1182_1_1}}\caption{Target}
\end{subfigure}\hfill\vfill
\makeatother
\caption{More qualitative comparison with other methods~\cite{dovenet, bargainet, s2am} on iHarmony4 Dataset. We mark the composite foreground mask as yellow region. S$^2$CRNet-S and  S$^2$CRNet-V  denote our method employing SqueezeNet and VGG16 backbone, respectively.}
\label{fig:supp_all}
\end{figure*}

%% file: main.bbl
\begin{thebibliography}{10}\itemsep=-1pt

\bibitem{rw_color2}
Vladimir Bychkovsky, Sylvain Paris, Eric Chan, and Fr{\'e}do Durand.
\newblock Learning photographic global tonal adjustment with a database of
  input/output image pairs.
\newblock In {\em CVPR}, pages 97--104. IEEE, 2011.

\bibitem{rw_color1}
Daniel Cohen-Or, Olga Sorkine, Ran Gal, Tommer Leyvand, and Ying-Qing Xu.
\newblock Color harmonization.
\newblock In {\em SIGGRAPH}, pages 624--630. 2006.

\bibitem{bargainet}
Wenyan Cong, Li Niu, Jianfu Zhang, Jing Liang, and Liqing Zhang.
\newblock {BargainNet}: Background-guided domain translation for image
  harmonization.
\newblock In {\em ICME}, 2021.

\bibitem{dovenet}
Wenyan Cong, Jianfu Zhang, Li Niu, Liu Liu, Zhixin Ling, Weiyuan Li, and Liqing
  Zhang.
\newblock Dovenet: Deep image harmonization via domain verification.
\newblock In {\em CVPR}, pages 8394--8403, 2020.

\bibitem{s2am}
Xiaodong Cun and Chi-Man Pun.
\newblock Improving the harmony of the composite image by spatial-separated
  attention module.
\newblock {\em TIP}, 29:4759--4771, 2020.

\bibitem{cun2020split}
Xiaodong Cun and Chi-Man Pun.
\newblock Split then refine: Stacked attention-guided resunets for blind single
  image visible watermark removal.
\newblock {\em AAAI}, 2021.

\bibitem{imagenet}
Jia Deng, Wei Dong, Richard Socher, Li-Jia Li, Kai Li, and Li Fei-Fei.
\newblock Imagenet: A large-scale hierarchical image database.
\newblock In {\em CVPR}, pages 248--255. Ieee, 2009.

\bibitem{Gharbi}
Micha{\"e}l Gharbi, Jiawen Chen, Jonathan~T Barron, Samuel~W Hasinoff, and
  Fr{\'e}do Durand.
\newblock {Deep bilateral learning for real-time image enhancement.}
\newblock {\em TOG}, 36(4):1--12, 2017.

\bibitem{gan}
Ian~J Goodfellow, Jean Pouget-Abadie, Mehdi Mirza, Bing Xu, David Warde-Farley,
  Sherjil Ozair, Aaron Courville, and Yoshua Bengio.
\newblock Generative adversarial networks.
\newblock {\em arXiv preprint arXiv:1406.2661}, 2014.

\bibitem{guo2020zero}
Chunle Guo, Chongyi Li, Jichang Guo, Chen~Change Loy, Junhui Hou, Sam Kwong,
  and Runmin Cong.
\newblock Zero-reference deep curve estimation for low-light image enhancement.
\newblock In {\em CVPR}, pages 1780--1789, 2020.

\bibitem{Guo_2021_CVPR}
Zonghui Guo, Haiyong Zheng, Yufeng Jiang, Zhaorui Gu, and Bing Zheng.
\newblock Intrinsic image harmonization.
\newblock In {\em CVPR}, pages 16367--16376, June 2021.

\bibitem{Hao2020bmcv}
Guoqing Hao, Satoshi Iizuka, and Kazuhiro Fukui.
\newblock Image harmonization with attention-based deep feature modulation.
\newblock In {\em BMVC}, 2020.

\bibitem{he2012guided}
Kaiming He, Jian Sun, and Xiaoou Tang.
\newblock Guided image filtering.
\newblock {\em TPAMI}, 35(6):1397--1409, 2012.

\bibitem{resnet}
Kaiming He, Xiangyu Zhang, Shaoqing Ren, and Jian Sun.
\newblock Deep residual learning for image recognition.
\newblock In {\em CVPR}, pages 770--778, 2016.

\bibitem{bvmr}
Amir Hertz, Sharon Fogel, Rana Hanocka, Raja Giryes, and Daniel Cohen-Or.
\newblock Blind visual motif removal from a single image.
\newblock In {\em CVPR}, pages 6858--6867, 2019.

\bibitem{howard2017mobilenets}
Andrew~G Howard, Menglong Zhu, Bo Chen, Dmitry Kalenichenko, Weijun Wang,
  Tobias Weyand, Marco Andreetto, and Hartwig Adam.
\newblock Mobilenets: Efficient convolutional neural networks for mobile vision
  applications.
\newblock {\em arXiv preprint arXiv:1704.04861}, 2017.

\bibitem{hu2018exposure}
Yuanming Hu, Hao He, Chenxi Xu, Baoyuan Wang, and Stephen Lin.
\newblock Exposure: A white-box photo post-processing framework.
\newblock {\em TOG}, 37(2):1--17, 2018.

\bibitem{iandola2016squeezenet}
Forrest~N Iandola, Song Han, Matthew~W Moskewicz, Khalid Ashraf, William~J
  Dally, and Kurt Keutzer.
\newblock Squeezenet: Alexnet-level accuracy with 50x fewer parameters and< 0.5
  mb model size.
\newblock {\em arXiv preprint arXiv:1602.07360}, 2016.

\bibitem{pix2pix}
Phillip Isola, Jun~Yan Zhu, Tinghui Zhou, and Alexei~A. Efros.
\newblock {Image-to-image translation with conditional adversarial networks}.
\newblock In {\em CVPR}, volume 2017-Janua, pages 5967--5976, nov 2017.

\bibitem{rw_gradient1}
Jiaya Jia, Jian Sun, Chi-Keung Tang, and Heung-Yeung Shum.
\newblock Drag-and-drop pasting.
\newblock {\em TOG}, 25(3):631--637, 2006.

\bibitem{alexnet}
Alex Krizhevsky, Ilya Sutskever, and Geoffrey~E Hinton.
\newblock Imagenet classification with deep convolutional neural networks.
\newblock {\em NIPS}, 25:1097--1105, 2012.

\bibitem{intro_color1}
Jean-Francois Lalonde and Alexei~A Efros.
\newblock Using color compatibility for assessing image realism.
\newblock In {\em ICCV}, pages 1--8. IEEE, 2007.

\bibitem{lee2019inserting}
Donghoon Lee, Tomas Pfister, and Ming-Hsuan Yang.
\newblock Inserting videos into videos.
\newblock In {\em CVPR}, pages 10061--10070, 2019.

\bibitem{mscoco}
Tsung-Yi Lin, Michael Maire, Serge Belongie, James Hays, Pietro Perona, Deva
  Ramanan, Piotr Doll{\'a}r, and C~Lawrence Zitnick.
\newblock Microsoft coco: Common objects in context.
\newblock In {\em ECCV}, pages 740--755, 2014.

\bibitem{Ling_2021_CVPR}
Jun Ling, Han Xue, Li Song, Rong Xie, and Xiao Gu.
\newblock Region-aware adaptive instance normalization for image harmonization.
\newblock In {\em CVPR}, pages 9361--9370, June 2021.

\bibitem{adamw}
Ilya Loshchilov and Frank Hutter.
\newblock Fixing weight decay regularization in adam.
\newblock 2018.

\bibitem{lsgan}
Xudong Mao, Qing Li, Haoran Xie, Raymond~YK Lau, Zhen Wang, and Stephen
  Paul~Smolley.
\newblock Least squares generative adversarial networks.
\newblock In {\em ICCV}, pages 2794--2802, 2017.

\bibitem{moran2021curl}
Sean Moran, Steven McDonagh, and Gregory Slabaugh.
\newblock Curl: Neural curve layers for global image enhancement.
\newblock In {\em 2020 25th International Conference on Pattern Recognition
  (ICPR)}, pages 9796--9803. IEEE, 2021.

\bibitem{pytorch}
Adam Paszke, Sam Gross, Soumith Chintala, Gregory Chanan, Edward Yang, Zachary
  DeVito, Zeming Lin, Alban Desmaison, Luca Antiga, and Adam Lerer.
\newblock Automatic differentiation in pytorch.
\newblock 2017.

\bibitem{rw_gradient2}
Patrick P{\'e}rez, Michel Gangnet, and Andrew Blake.
\newblock Poisson image editing.
\newblock In {\em SIGGRAPH}, pages 313--318. 2003.

\bibitem{intro_color2}
Erik Reinhard, Michael Adhikhmin, Bruce Gooch, and Peter Shirley.
\newblock Color transfer between images.
\newblock {\em CG\&A}, 21(5):34--41, 2001.

\bibitem{unet}
Olaf Ronneberger, Philipp Fischer, and Thomas Brox.
\newblock U-net: Convolutional networks for biomedical image segmentation.
\newblock In {\em MICCAI}, pages 234--241. Springer, 2015.

\bibitem{simonyan2014very}
Karen Simonyan and Andrew Zisserman.
\newblock Very deep convolutional networks for large-scale image recognition.
\newblock {\em arXiv preprint arXiv:1409.1556}, 2014.

\bibitem{sofiiuk2021foreground}
Konstantin Sofiiuk, Polina Popenova, and Anton Konushin.
\newblock Foreground-aware semantic representations for image harmonization.
\newblock In {\em WACV}, pages 1620--1629, 2021.

\bibitem{intro_texture}
Kalyan Sunkavalli, Micah~K Johnson, Wojciech Matusik, and Hanspeter Pfister.
\newblock Multi-scale image harmonization.
\newblock {\em TOG}, 29(4):1--10, 2010.

\bibitem{DIH_Tsai}
Yi-Hsuan Tsai, Xiaohui Shen, Zhe Lin, Kalyan Sunkavalli, Xin Lu, and Ming-Hsuan
  Yang.
\newblock Deep image harmonization.
\newblock In {\em CVPR}, 2017.

\bibitem{intro_illumination}
Su Xue, Aseem Agarwala, Julie Dorsey, and Holly Rushmeier.
\newblock Understanding and improving the realism of image composites.
\newblock {\em TOG}, 31(4):1--10, 2012.

\bibitem{3dlut}
Hui Zeng, Jianrui Cai, Lida Li, Zisheng Cao, and Lei Zhang.
\newblock Learning image-adaptive 3d lookup tables for high performance photo
  enhancement in real-time.
\newblock {\em TPAMI}, 2020.

\bibitem{zhou2018semantic}
Bolei Zhou, Hang Zhao, Xavier Puig, Tete Xiao, Sanja Fidler, Adela Barriuso,
  and Antonio Torralba.
\newblock Semantic understanding of scenes through the ade20k dataset.
\newblock {\em IJCV}, 2018.

\bibitem{relighting}
Hao Zhou, Sunil Hadap, Kalyan Sunkavalli, and David~W. Jacobs.
\newblock Deep single portrait image relighting.
\newblock In {\em ICCV}, 2019.

\bibitem{zhou2018learning}
Peng Zhou, Xintong Han, Vlad~I Morariu, and Larry~S Davis.
\newblock Learning rich features for image manipulation detection.
\newblock In {\em CVPR}, pages 1053--1061, 2018.

\bibitem{zhu2015learning}
Jun-Yan Zhu, Philipp Krahenbuhl, Eli Shechtman, and Alexei~A Efros.
\newblock Learning a discriminative model for the perception of realism in
  composite images.
\newblock In {\em ICCV}, pages 3943--3951, 2015.

\bibitem{cyclegan}
Jun-Yan Zhu, Taesung Park, Phillip Isola, and Alexei~A Efros.
\newblock {Unpaired Image-to-Image Translation using Cycle-Consistent
  Adversarial Networks}.
\newblock {\em ICCV}, Mar. 2017.

\end{thebibliography}
